\documentclass[12pt]{article}
\usepackage{amsmath}
\usepackage{amssymb}
\usepackage{graphicx,psfrag,epsf}
\usepackage{enumerate}
\usepackage[numbers,sort&compress]{natbib}
\usepackage[colorlinks=true, citecolor=blue, linkcolor=blue, urlcolor=blue]{hyperref}
\usepackage{xcolor}
\newcommand{\blind}{0}
\usepackage{float}
\usepackage{booktabs}
\usepackage{authblk} % nicely formatted authors + affiliations
\usepackage{longtable}
\usepackage{tabularx}
\usepackage{caption}
\usepackage{geometry}

\begin{document}

\def\spacingset#1{\renewcommand{\baselinestretch}%
{#1}\small\normalsize} \spacingset{1}

%%%%%%%%%%%%%%%%%%%%%%%%%%%%%%%%%%%%%%%%%%%%%%%%%%%%%%%%%%%%%%%%%%%%%%%%%%%%%%

% Replace your entire \if0\blind {...} block with this:
\if0\blind
{
  \title{\bf Predicting Mycotoxin Contamination in Irish Oats Using Deep and Transfer Learning}

  \author[1]{Alan Inglis}
  \author[2]{Fiona Doohan}
  \author[2]{Subramani Natarajan}
  \author[3]{Breige McNulty}
  \author[4]{Chris Elliott}
  \author[3,4]{Anne Nugent}
  \author[4]{Julie Meneely}
  \author[4,5]{Brett Greer}
  \author[6]{Stephen Kildea}
  \author[6]{Diana Bucur}
  \author[7]{Martin Danaher}
  \author[7]{Melissa Di Rocco}
  \author[8]{Lisa Black}
  \author[8]{Adam Gauley}
  \author[8]{Naoise McKenna}
  \author[9]{Andrew Parnell}

  \affil[1]{Hamilton Institute, Maynooth University, Ireland}
  \affil[2]{UCD School of Biology and Environmental Science, University College Dublin, Ireland}
  \affil[3]{UCD School of Agriculture and Food Science \& The Institute of Food and Health, University College Dublin, Ireland}
  \affil[4]{Institute for Global Food Security, School of Biological Sciences, Queen's University Belfast, Belfast, UK}
  \affil[5]{Chemical Surveillance Branch, Veterinary Sciences Division, Agri-Food and Biosciences Institute (AFBI), Belfast, UK}
  \affil[6]{Crops Research Centre, Teagasc, Oak Park, Ireland}
  \affil[7]{Food Safety Department, Teagasc Food Research Centre, Ashtown, Ireland}
  \affil[8]{Agri-Food and Biosciences Institute (AFBI), Belfast, UK}
  \affil[9]{UCD School of Mathematics and Statistics, University College Dublin, Ireland}

  \date{} % remove date
  \maketitle
} \fi

\bigskip
\begin{abstract}
Mycotoxin contamination poses a significant risk to cereal crop quality, food safety, and agricultural productivity. Accurate prediction of mycotoxin levels can support early intervention strategies and reduce economic losses. This study investigates the use of neural networks and transfer learning models to predict mycotoxin contamination in Irish oat crops as a multi-response prediction task. Our dataset comprises oat samples collected in Ireland, containing a mix of environmental, agronomic, and geographical predictors. Five modelling approaches were evaluated: a baseline multilayer perceptron (MLP), an MLP with pre-training, and three transfer learning models; TabPFN, TabNet, and FT-Transformer. Model performance was evaluated using regression (RMSE, $R^2$) and classification (AUC, F1) metrics, with results reported per toxin and on average. Additionally, permutation-based variable importance analysis was conducted to identify the most influential predictors across both prediction tasks. The transfer learning approach TabPFN provided the overall best performance, followed by the baseline MLP. Our variable importance analysis revealed that weather history patterns in the 90-day pre-harvest period were the most important predictors, alongside seed moisture content.
\end{abstract}

\noindent%

\spacingset{1.45}

\section{Introduction}
\label{sec:intro}

Mycotoxins are toxic secondary metabolites produced by certain species of fungi, most notably those in the genera \textit{Fusarium}, \textit{Aspergillus}, and \textit{Penicillium}. These compounds frequently contaminate cereal crops, including oats, wheat, barley, and corn, both in the field and during storage \citep{WHO}. The mycotoxins of greatest concern in European cereals include deoxynivalenol, zearalenone, aflatoxins, fumonisins, ochratoxin A, and trichothecenes such as T-2 and HT-2 toxins \citep{eskola2020worldwide}. Exposure to mycotoxins poses serious risks to human and animal health, with acute effects including nausea, vomiting, and feed refusal in livestock \citep{mavrommatis2021impact,moon2002vomitoxin}, while chronic exposure in humans is linked to carcinogenicity, hepatotoxicity, and mortality \citep{marroquin2014mycotoxins,liu2010global}. To protect public health, regulatory frameworks in Europe set maximum levels for mycotoxins in food and feed \citep{eskola2020worldwide}.

The economic impact of mycotoxin contamination is substantial. Globally, surveillance, compliance, and mitigation costs are estimated in the billions of euros annually \citep{wu2015global}, with contamination affecting 60–80\% of crops and approximately 20\% exceeding legal limits in some assessments \citep{eskola2020worldwide}. Trade disruptions further amplify these losses \citep{alshannaq2017occurrence, marroquin2014mycotoxins}. In Europe, between 2010 and 2019, an estimated 75 million tonnes of wheat (about 5\% intended for human consumption) exceeded deoxynivalenol limits, resulting in downgrading to feed status and losses of approximately EUR 3 billion \citep{johns2022emerging}. Aflatoxin contamination contributed to the downgrading of an additional 4.2\% of wheat between 2010 and 2020, with potential losses of about EUR 2.5 billion \citep{latham2023diverse}.

Mycotoxin occurrence varies significantly across geographic regions and seasons, reflecting local weather patterns and agronomic conditions \citep{logrieco2021perspectives,leggieri2020impact}. Since 2012, several reports indicate increased incidence across Europe, with climate change likely contributing through shifts in temperature and moisture regimes that favour fungal growth \citep{zingales2022climate,medina2017climate}. In temperate maritime climates such as Ireland, these environmental conditions are particularly conducive to fungal proliferation and mycotoxin formation in small grain cereals \citep{de2021natural}, making predictive approaches especially valuable for risk management in such settings.

While laboratory-based analytical techniques remain the gold standard for confirming mycotoxin contamination in cereal crops, their role is primarily diagnostic and typically occurs after contamination has already taken place. Methods such as high-performance liquid chromatography (HPLC) and liquid chromatography coupled with tandem mass spectrometry (LC-MS/MS) provide high accuracy and specificity in mycotoxin detection \citep{anfossi2016mycotoxin, maragos2004emerging}. Enzyme-linked immunosorbent assays (ELISA) are simpler and faster alternatives that enable rapid screening with minimal equipment and training requirements. However, ELISAs are typically semi-quantitative and specific to individual mycotoxins (e.g. deoxynivalenol, zearalenone) or classes of mycotoxins (e.g. aflatoxins), requiring multiple kits to cover the full range of regulated toxins. Any samples indicating potential non-compliance generally require confirmatory analysis using LC-MS/MS. Despite these trade-offs, all such laboratory-based methods can be time-consuming and may not be practical for large-scale, on-site, or real-time monitoring, particularly within globally distributed supply chains where rapid decision-making is essential \citep{soares2018advances, renaud2019mycotoxin}. Consequently, there is growing interest in complementary predictive approaches that can support early warning and risk mitigation. These approaches, including machine learning-based decision support systems, do not replace laboratory testing but instead provide actionable forecasts that help prioritise sampling, optimise resource allocation, and reduce contamination risk before laboratory confirmation is available.

Machine learning (ML) methods have emerged as promising tools for predicting mycotoxin contamination in cereal crops. These approaches leverage diverse data sources, including weather patterns, agronomic practices, soil characteristics, and remote sensing imagery, to model the complex interactions that drive mycotoxin production \citep{mu2024making}. Various algorithms have demonstrated success in this domain. For example, Random Forest models \citep{breiman2001random} have been combined with near-infrared spectroscopy to classify mycotoxins in corn and oats \citep{ghilardelli2022preliminary, teixido2023quantification}, while gradient boosting methods \citep{friedman2001greedy} have been applied to aflatoxin monitoring in peanuts and soybeans \citep{wang2022designing} and fumonisin detection in corn \citep{chavez2022single}. More recently, deep learning architectures, particularly artificial neural networks (ANNs), have shown capacity to capture complex non-linear relationships in high-dimensional agricultural datasets, with applications ranging from aflatoxin and fumonisin detection in Italian corn \citep{camardo2021machine} to deoxynivalenol and nivalenol analysis in Polish winter wheat \citep{niedbala2020application}. For a comprehensive review of ML applications in mycotoxin detection, see \citet{inglis2024machine}.

While ML approaches offer significant advantages over traditional statistical models, including the ability to integrate heterogeneous datasets, handle numerous predictor variables, and identify subtle patterns, there remains scope to explore newer modelling approaches that might further improve predictive performance \citep{inglis2024machine}. Transfer learning (TL) addresses this limitation by enabling models to leverage knowledge acquired from large, diverse source datasets and apply it to smaller, domain-specific target datasets \citep{weiss2016survey}. The fundamental principle underlying TL is that models trained on a dataset with large training data can be applied to a task with limited data availability, thereby reducing the need for extensive training datasets in the target application \citep{lu2015transfer}. This capability is particularly valuable in agricultural contexts where obtaining large, well-annotated datasets can be challenging and expensive. TL has achieved notable success in computer vision \citep{li2021benchmarking, li2020transfer, gopalakrishnan2017deep} and natural language processing \citep{ruder2019transfer, wang2015transfer} and is gaining traction in agricultural applications \citep{hossen2025transfer, ruder2019transfer}. By reusing learned representations from broader contexts, TL can potentially improve predictive performance and generalisation in mycotoxin prediction without requiring extensive labelled training datasets.

Although TL applications in mycotoxin prediction remain limited, several studies demonstrate its potential. For example, \citet{kilicc2022novel} developed a non-invasive classification system for aflatoxin contamination in dried figs, providing an alternative to conventional manual sorting with UV light. \citet{deng2025enhancing} successfully demonstrated cross-commodity prediction between wheat zearalenone and peanut aflatoxin B1 using TL-enhanced chemometric modelling. Additionally, \citet{qiu2019detection} applied TL to detect \textit{Fusarium} head blight in wheat. These examples suggest that TL can support more generalisable agricultural risk monitoring models, though multi-toxin, multi-crop applications remain under-explored.

Our study addresses these gaps by applying neural networks (NNs) and TL to predict mycotoxin occurrence across Ireland using data from approximately 100 locations. The dataset includes 24 different mycotoxins and a wide range of predictors including soil chemistry, crop management practices, cultivar information, and time-resolved weather patterns. In our NN framework, we adopt a multi-task learning approach that simultaneously predicts both mycotoxin presence (classification) and concentration (regression). Not all TL architectures evaluated support multitask training, so for those models classification and regression tasks were implemented separately. The models considered include a standard feedforward multilayer perceptron (MLP), a pre-trained MLP initialised via an autoencoder, and three transfer learning architectures: TabPFN, TabNet, and FT-Transformer. To address the challenge of missing values in mycotoxin measurements, we implement custom neural network loss functions that mask missing outcomes, allowing training across multiple responses without discarding incomplete cases. While NNs and TL have previously been explored for mycotoxin detection, this work represents the first study in Ireland to apply these approaches to mycotoxin prediction at this scale, and, to our knowledge, the first to employ a masked loss framework for multitask prediction in this context. The primary objectives of this work are to assess the predictive performance of ML and TL models across multiple mycotoxin responses, to compare the effectiveness of different modelling approaches for multitask prediction, and to evaluate the potential of TL methods for improving contaminant risk prediction in cereal production systems.

This paper is structured as follows: Section \ref{sec:methods} details the methods used throughout the work. This includes data collection and preprocessing steps, exploratory data analysis, and model architectures (that is, neural network architectures, custom loss functions, and TL strategies employed). Section \ref{sec:results} describes the results which include overall model and per-toxin performances, as well as variable importance. Finally, in Section \ref{sec:conclusion}, we provide some concluding remarks.

% in this paper

\section{Materials and Methods}
\label{sec:methods}

\subsection{Data Sources}
\label{subsec:data}

\begin{figure}[H]
\centering
\includegraphics[width=0.5\textwidth]{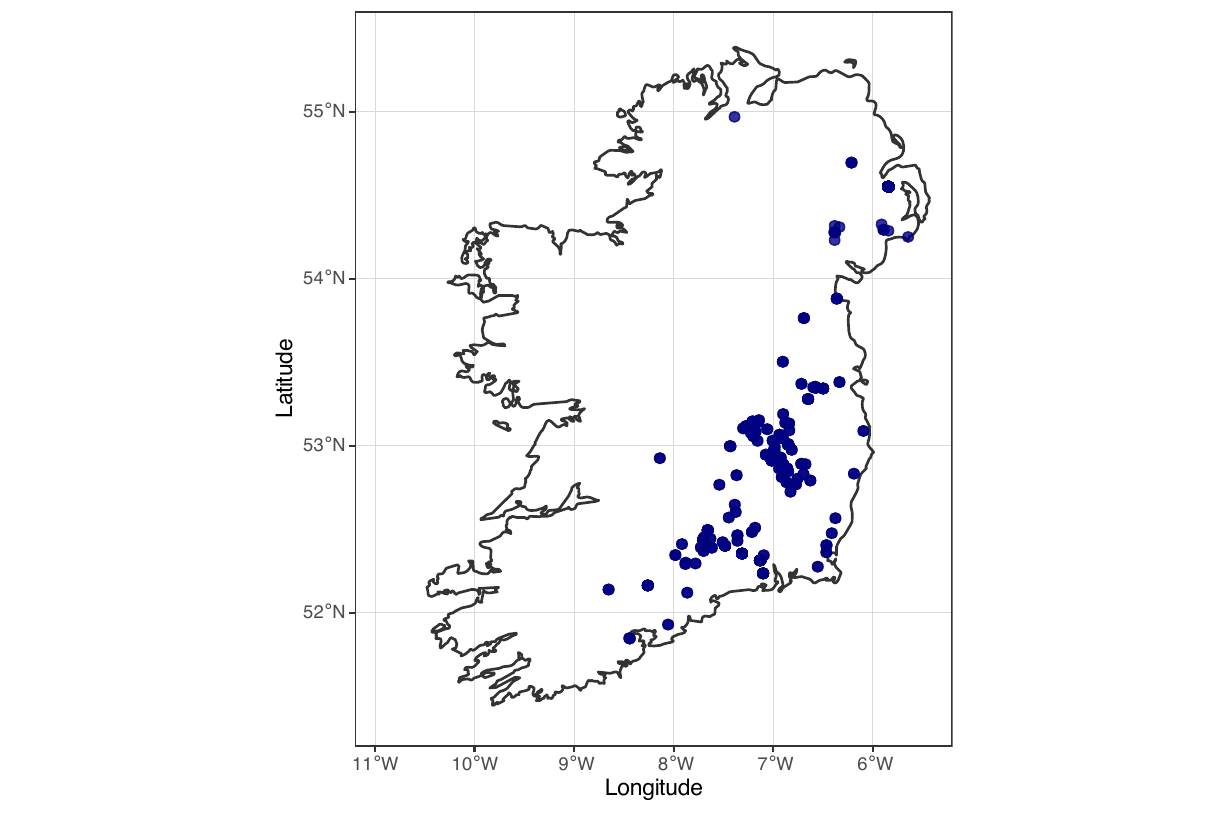}
\caption{Sampling site locations.}
\label{fig:site}
\end{figure}

The dataset analysed in this study was compiled through the \textit{Mycotox-I} project\footnote{\url{https://www.mycotoxi.com/}}, a multi-institutional collaboration across Ireland (see Acknowledgments and Funding for full details of participating institutions). Data were collected from approximately 100 sampling locations across the island of Ireland between 2022 and 2023, spanning both institutional research sites and commercial farms. The majority of sites were situated in the eastern counties (for example, Kildare, Laois, Carlow, Kilkenny, Wexford, and Tipperary), with additional sites extending southwards into Cork and Waterford and northwards into Louth, Meath, and Northern Ireland (Antrim, Armagh, Down, Derry). This distribution provided a north–south gradient across the island while concentrating on the main oats-growing regions (see Figure \ref{fig:site}).

\begin{figure}[H]
\centering
\includegraphics[width=0.8\textwidth]{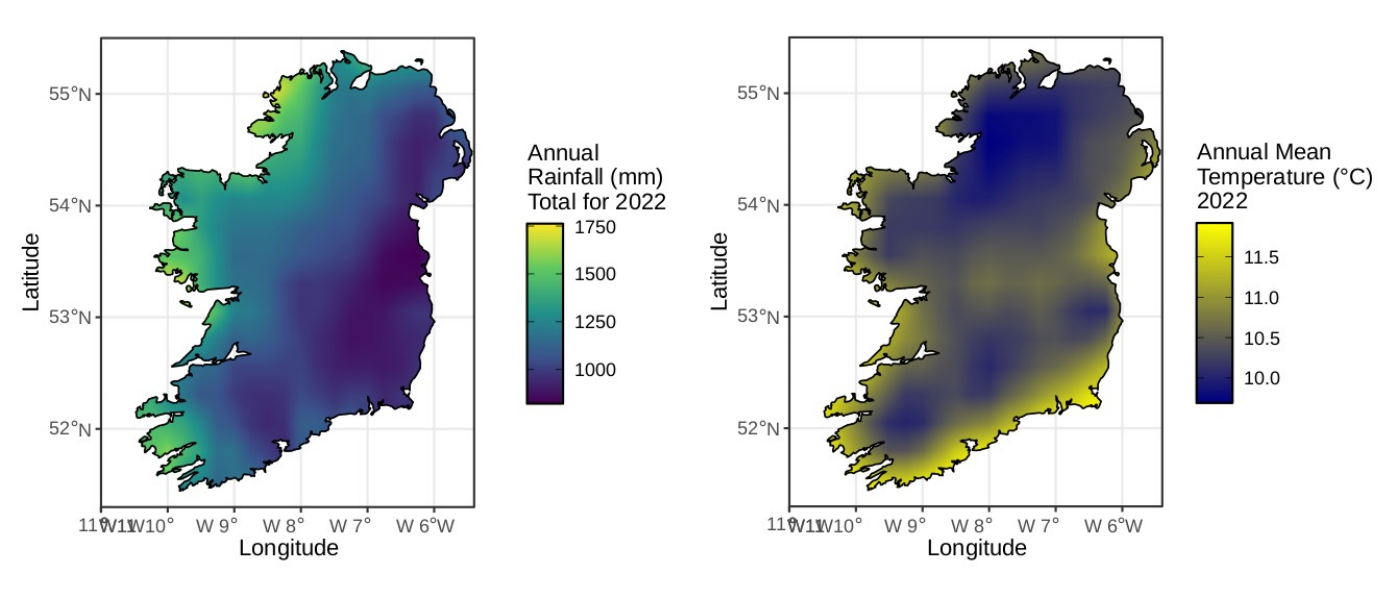}
\caption{Total annual rainfall (mm) and annual mean temperature ($^{\circ}$C) for 2022.}
\label{fig:weather}
\end{figure}

Daily weather data (precipitation, air temperature, and dew point) were sourced from ERA5 reanalysis NetCDF files \citep{hersbach2020era5}.  Most sampling sites were located in relatively low-rainfall areas of the east, with some positioned in regions characterised by slightly warmer microclimates in the south-east and south. 
Figure \ref{fig:weather} shows the total annual  rainfall in millimetres and the mean annual temperature in $^\circ$C for the year 2022. These maps show the national gradients, highlighting wetter western regions and cooler northern areas compared with less wet eastern regions and warmer southern areas. 

Samples were originally analysed for 33 mycotoxins and related metabolites, including three ergot alkaloids. For the purposes of this study, only 24 analytes were retained for modelling. The remaining analytes were excluded due to near-zero concentrations or extensive missing data, which precluded reliable statistical analysis (see Supplementary Materials for a complete list). Laboratory analysis was conducted using Ultra-High Performance Liquid Chromatography (UHPLC)-MS/MS (see section 2.2.3 for details). Mycotoxin concentrations recorded as below the limit of quantification (LOQ) were assigned a value of zero within the database, following the rationale that concentrations that are not quantifiable represent negligible contamination levels. This approach is consistent with previous studies that have adopted similar methodologies for handling below the LOQ mycotoxin data \citep{ec2023commission, wang2022regional, pallares2019mycotoxin}. Missing values were retained as \texttt{NA} where mycotoxins were not measured at specific locations, allowing these to be handled appropriately within the modelling framework.

The predictor dataset included comprehensive environmental, agronomic, and crop-specific variables, totalling 40 agronomic predictors, three weather variables, and approximately 300 observations overall. Environmental data comprised daily weather records (temperature, precipitation, relative humidity) for the 90 days preceding harvest, reflecting the critical period for mycotoxin development. Soil characteristics included pH and concentrations of key nutrients (phosphorus, potassium, manganese, copper, zinc), along with soil type classifications. Agronomic variables captured detailed management practices including sowing and harvest dates, seed rates, and comprehensive records of fertiliser, fungicide, insecticide, and growth regulator products with their respective application timings and dosages. Crop rotation history was recorded for up to five previous years. Crop-specific descriptors included cereal species, variety, sowing ideotype (spring/winter), establishment and cropping systems. Additional variables included yield, moisture content, green leaf area, foliar disease incidence, and geospatial information (latitude, longitude, county, elevation). The response set comprised 24 distinct mycotoxin measurements. A complete list of all predictor variables is provided in the Supplementary Materials.

\subsection{Oat Sample Analysis}

\subsubsection{Sample Preparation and UHPLC-MS/MS Analysis}

Reagents used were of analytical grade. Mycotoxin and ergot alkaloid standards, including stable isotopically labelled internal standards, were obtained from Romer Laboratory (Tulln, Austria) and University of Natural Resources and Life Sciences (BOKU, Vienna). An 8-point calibration curve was prepared with analytes ranging from 0.5 to 500 \textmu g/kg, covering the 24 analytes retained for modelling. All samples were extracted according to the dilute-and-shoot protocol as outlined by Sulyok et al. \citep{sulyok2020validation} with minor modifications. 

Quantitative analysis was performed using a SCIEX Exion LC™ AD UHPLC coupled with a SCIEX 5500+ QTrap triple quadrupole mass spectrometer (MS/MS). Chromatographic separation was achieved on a Gemini C18-column (100 $\times$ 4.6 mm, 5 \textmu m) using a binary gradient of methanol/water/acetic acid containing 5 mM ammonium acetate buffer. The system operated in scheduled multiple reaction monitoring mode in both positive and negative polarities. Mass spectrometric parameters, including monitored transitions, declustering potentials, collision exit cell potentials, and collision energies, are shown in the Supplementary Materials.

\subsubsection{Method Validation}

Method validation was performed in accordance with the SANTE/11312/2021 guidelines \citep{pihlstrom2021analytical}, adapting protocols described by Sulyok et al. \citep{sulyok2020validation} and Steiner et al. \citep{steiner2020realizing}. To account for matrix effects and lot-to-lot variation, five distinct oat varieties were spiked at two concentration levels (low and high, see Supplementary Materials) and analysed over three days. Performance metrics, including within-laboratory reproducibility ($RSD_{WLR}$), repeatability, extraction efficiency, and recovery were assessed. The limit of quantification (LOQ) and limit of detection (LOD) were determined following Eurachem guidelines \citep{bertil2014fitness}. The analytical method is accredited to ISO/IEC 17025 standards.

\subsection{Exploratory Data Analysis}
\label{subsec:eda}

\begin{figure}[H]
\centering
\includegraphics[width=1\textwidth,trim={0 .7cm 0 0},clip]{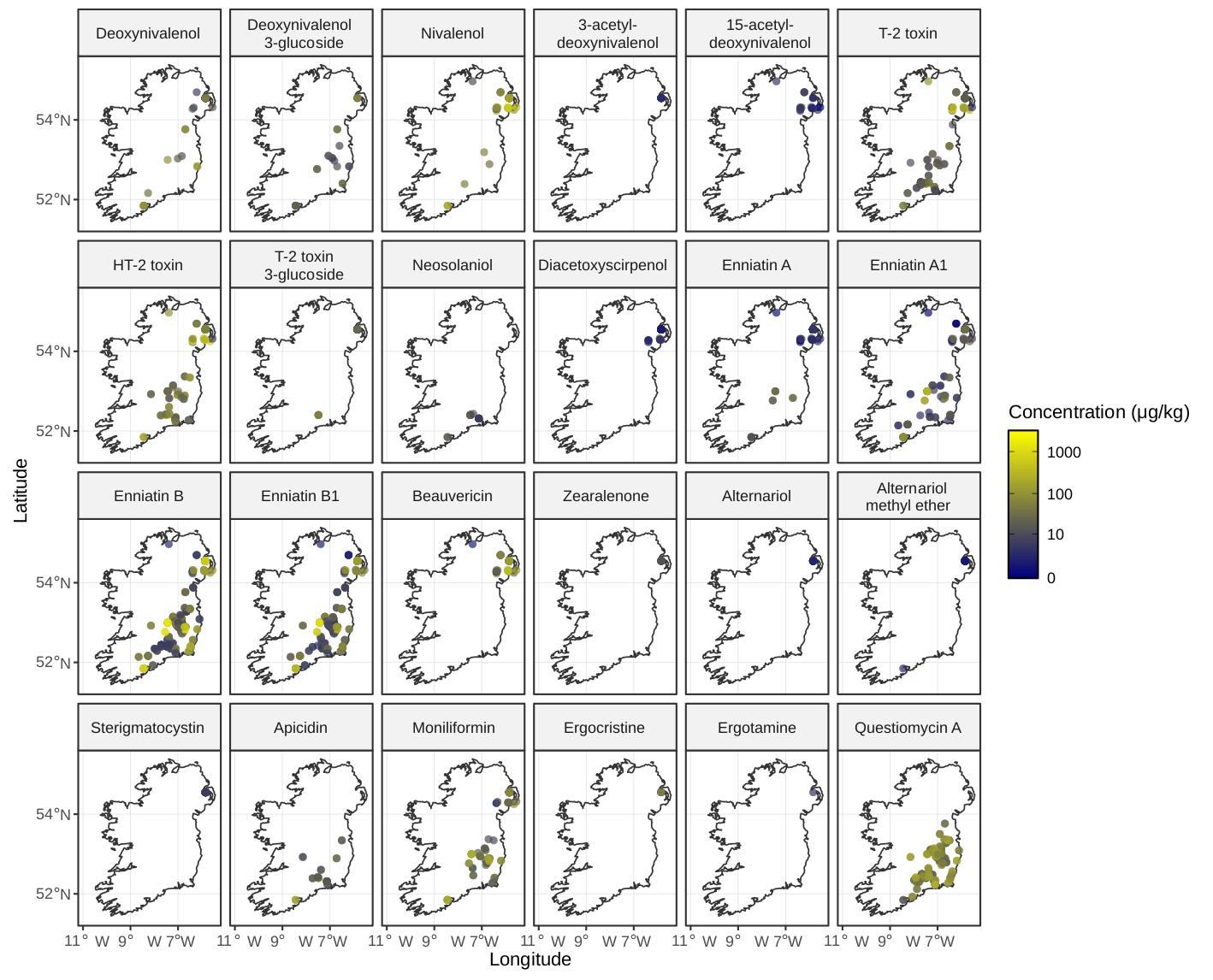}
\caption{Spatial distribution of contamination concentrations by individual mycotoxin.}
\label{fig:toxin_map}
\end{figure}

To characterise the contamination patterns and overall structure of the dataset, an exploratory data analysis was conducted across all 24 measured mycotoxins and associated predictor variables. Figure \ref{fig:toxin_map} displays the geographic distribution of each mycotoxin contamination across sampling locations. The colour intensity of each point reflects the total concentration of a particular mycotoxin at that location. Sampling effort was concentrated primarily in eastern regions, spanning from north to south. Several sites in both the northern and southern extremes of this zone show comparatively high contamination levels, with concentrations exceeding the upper quartile of the distribution. Figure \ref{fig:toxin_map} highlights the spatial diversity of toxin occurrence, showing that some compounds are widely distributed (for example, T-2 toxin and HT-2 toxin), while others appear only sporadically (for example, neosolaniol or diacetoxyscirpenol). A subtle north–south gradient can also be observed for certain toxins, suggesting potential regional differences in environmental or agronomic drivers that may warrant further investigation.

\begin{figure}[H]
\centering
\includegraphics[width=0.8\textwidth]{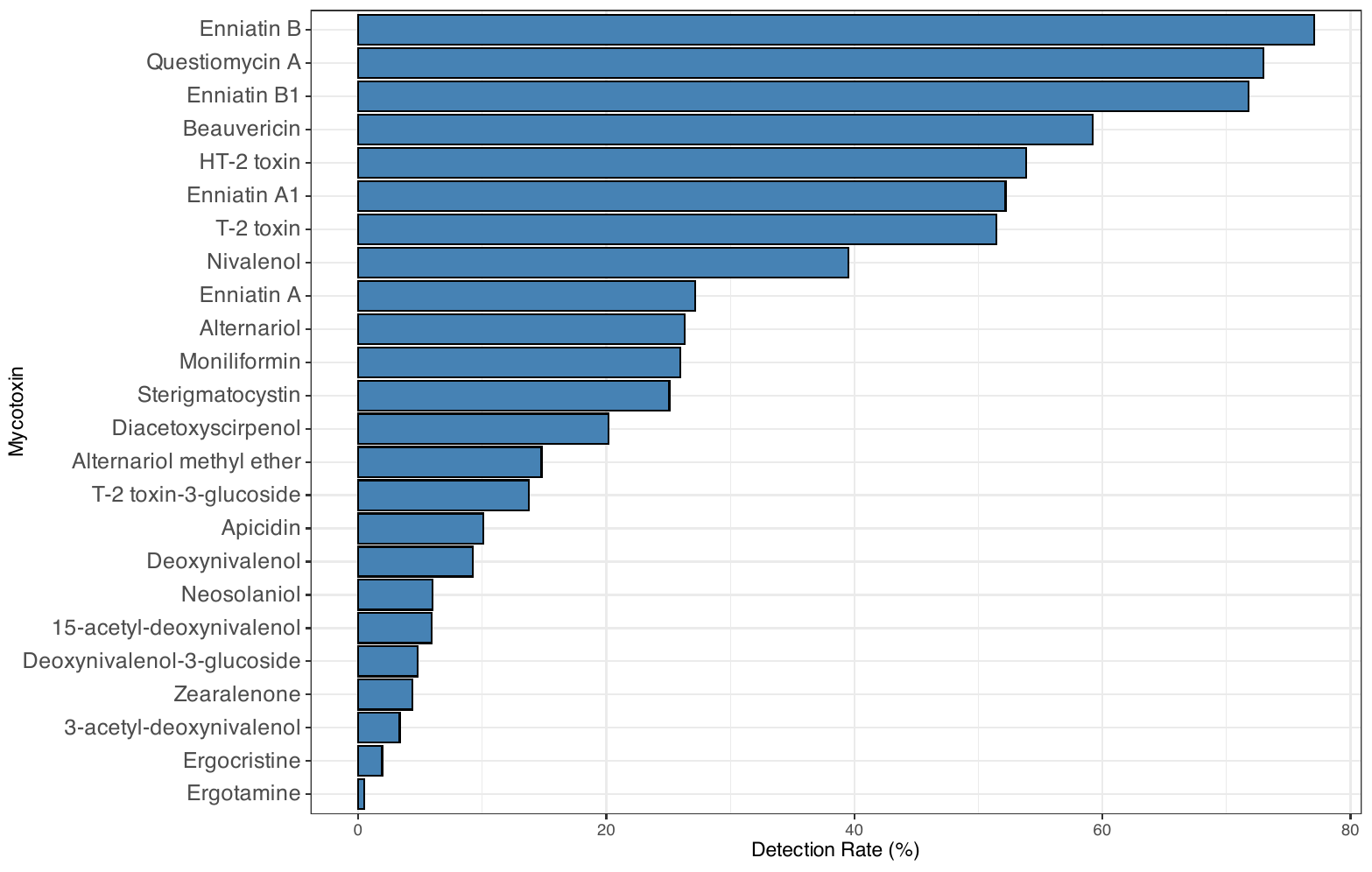}
\caption{Detection rates for all mycotoxins measured, showing the percentage of samples in which each compound was detected above the limit of quantification.}
\label{fig:detection_rates}
\end{figure}

Figure \ref{fig:detection_rates} shows the detection frequencies of all analysed mycotoxins. Detection rates varied considerably across compounds. Enniatins were the most frequently detected, with ENN~B, ENN~B1, and ENN~A1 occurring in over half of all samples (77\%, 72\%, and 52\%, respectively). Questiomycin~A was also common, detected in 73\% of samples. In contrast, several compounds were detected infrequently, including the ergot alkaloids ergotamine (0.5\%) and ergocristine (1.9\%), and 3-acetyl-deoxynivalenol (3.3\%). Maximum concentrations also varied widely, from 4.5 \textmu g/kg for ergotamine to 3,127 \textmu g/kg for HT-2 toxin, followed by ENN B at 2,852 \textmu g/kg. Summary statistics for all toxins are provided in the Supplementary Materials.

\begin{figure}[H]
\centering
\includegraphics[width=0.8\textwidth]{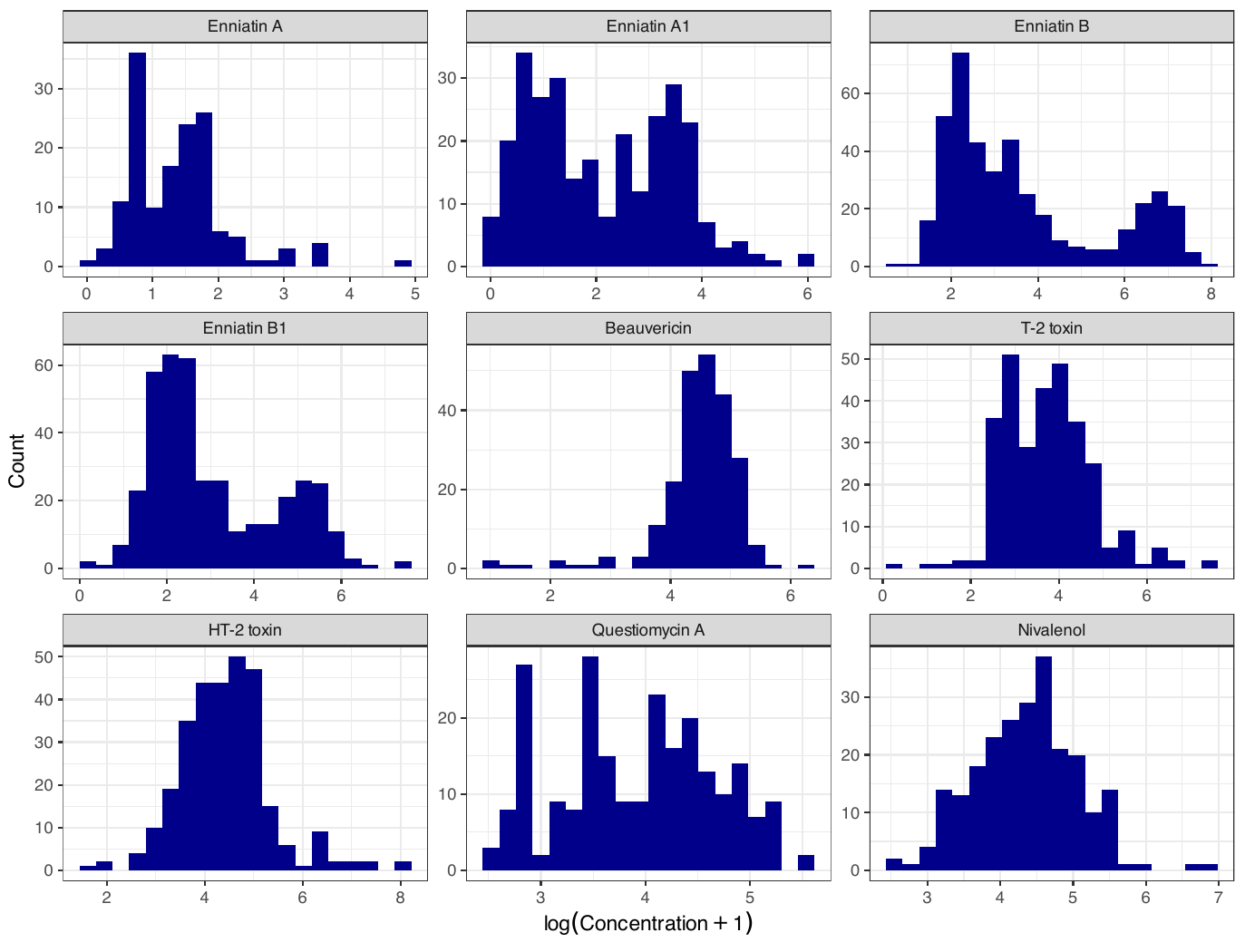}
\caption{Concentration distributions for the nine most frequently detected mycotoxins. Histograms show log$(x+1)$ transformed concentrations for samples where each mycotoxin was detected above the limit of quantification.}
\label{fig:concentration_distributions}
\end{figure}

The concentration distributions for the nine most frequently detected mycotoxins are shown in Figure \ref{fig:concentration_distributions}. These histograms display the log-transformed concentration values for samples where each compound was detected above the limit of quantification. Most distributions are right-skewed, reflecting that high concentration events occur infrequently. Beauvericin shows a slight left-skewed pattern, with concentrations clustering towards higher values. Both T-2 and HT-2 toxins display relatively symmetric distributions centred around intermediate concentration levels. ENN B and ENN B1 show evidence of bimodality, characterised by a dominant peak at lower concentrations and a smaller secondary peak at higher levels. Questiomycin A and nivalenol also appear broadly symmetric but with greater spread, indicating wider variability across samples.

\begin{figure}[H]
\centering
\includegraphics[width=0.8\textwidth]{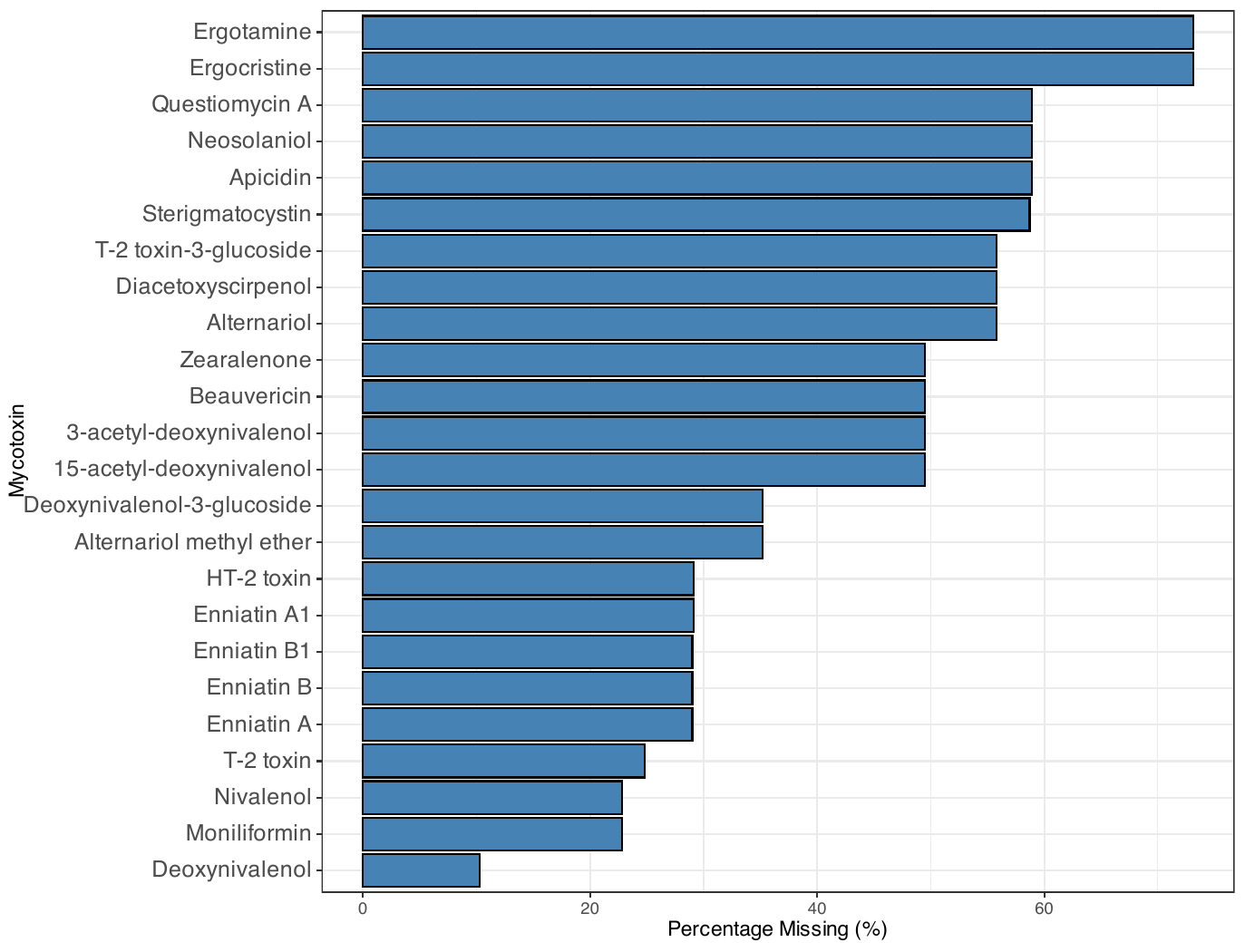}
\caption{Percentage of missing data for each mycotoxin across all sampling locations, showing the completeness of measurements for each compound.}
\label{fig:missing_data}
\end{figure}

Missing data patterns across the mycotoxin measurements are shown in Figure \ref{fig:missing_data}, reflecting instances where specific compounds were not measured at particular sampling locations. Missingness varied substantially across toxins. The highest proportions of missing data were observed for the ergot alkaloids, ergocristine and ergotamine (both 73.1\%), followed by neosolaniol, apicidin, questiomycin A, and sterigmatocystin (approximately 59\%). Moderate levels of missingness (around 50\%) were found for compounds such as
diacetoxyscirpenol, T-2 toxin 3 glucoside, beauvericin, and several type B trichothecenes, including 3-acetyl-deoxynivalenol and 15-acetyl-deoxynivalenol. In contrast, enniatins and HT-2 toxin showed comparatively higher data completeness, with missing values ranging from 24--29\%. The lowest levels of missing data were recorded for nivalenol (22.8\%), moniliformin (22.8\%), and deoxynivalenol (10.3\%). These differences primarily reflect variations in analytical coverage across compounds and sampling campaigns.

\subsection{Data Preparation}
\label{subsec:preprocessing}

The raw dataset was pre-processed to handle missing values and prepare features for modelling. Concentrations below the LOQ  were imputed as zero and, where soil pH was reported as a range, the midpoint was taken. All continuous response variables represented the concentration of a given mycotoxin in \textmu g/kg, with values log-transformed using log$(x+1)$ to address the skewed distribution of concentrations. Corresponding binary response variables were derived by coding concentrations greater than zero as 1 (indicating mycotoxin presence) and zero as 0 (indicating absence). To capture seasonality, dates for sowing, input applications, and harvest were encoded as sine–cosine transformations of the day of year: 
\begin{equation*}
Day_{sin} = \sin\left(\tfrac{2\pi d}{365}\right), \quad Day_{cos} =\cos\left(\tfrac{2\pi d}{365}\right),
\end{equation*}
where $d$ is the calendar day. This representation preserves the cyclical nature of dates (for example, $31^{st}$ December and $1^{st}$ January are adjacent).

From the temperature and dew point data obtained from ERA5 reanalysis NetCDF, the daily relative humidity was calculated using the August-Roche-Magnus approximation \citep{alduchov1996improved}:
\begin{equation*}
RH = 100 \times \frac{\exp\left(\tfrac{17.625 T_d}{243.04 + T_d}\right)}
{\exp\left(\tfrac{17.625 T}{243.04 + T}\right)},
\end{equation*}
where $T_d$ is dew point temperature (°C) and $T$ is air temperature (°C). For each sample, weather variables were extracted by matching the latitude and longitude coordinates of the sampling farms to the nearest ERA5 grid points. The precipitation, temperature, and relative humidity data from these matched grid cells for each day, up to 90 days,  preceding harvest were then extracted to serve as lagged predictors.

Categorical predictors were one-hot encoded, and continuous predictors were normalised to improve model convergence. Missing predictor values were addressed through multiple imputation using the \texttt{mice} R package \citep{mice}, with logistic regression for binary features and random forests for other numeric and factor variables. Any remaining sparse columns ($>95\%$ missing) or those with missing values post-imputation were removed. Missingness in the response variables, as shown in Figure \ref{fig:missing_data}, was handled during model training via a custom masked loss function (see Section \ref{subsec:baseline_fnn}), which allowed for the retention of partially observed samples.

The final dataset was split into training (80\%) and test (20\%) sets using a procedure to preserve the distribution of key predictors and mycotoxin occurrences, with all samples from the same location and year combination assigned to the same split. This was done so as to use the same training/test sets in each model.

\subsection{Model Architectures}
\label{subsec:models}

To predict multi-response concentrations and presence–absence, we evaluated five models. These include; two feedforward multilayer perceptrons (MLPs) and three specialised tabular models. The baseline NN was a custom multi-head MLP implemented in \texttt{R} (v4.4.2, \citep{R}) using \texttt{keras3} \citep{keras} with a TensorFlow backend \citep{tensorflow}. The second MLP was identical in architecture but initialised via autoencoder pre-training (pre-trained MLP). The remaining models were FT-Transformer \citep{gorishniy2021revisiting}, TabNet \citep{arik2021tabnet}, and TabPFN \citep{hollmann2022tabpfn}. These tabular learning models were implemented in \texttt{Python}, with their specific architectures detailed in their respective sections below.

Table \ref{tab:model_summary} provides an overview of the key characteristics and implementation constraints for each model architecture. Custom loss functions were used in the baseline model to handle missing response values and prevent data loss during training. However, it should be noted that some tabular learning models do not natively support custom output heads or loss functions. In such cases, the default architecture was used, with adaptations discussed in detail within each respective model section.

\begin{table}[H]
\centering
\caption{Summary of Model Architectures and Implementation Characteristics}
\label{tab:model_summary}
\begin{tabular}{l|c|c|c|c|c}
\toprule
\textbf{Model} & \textbf{Custom} & \textbf{Custom} & \textbf{Multi-task} & \textbf{Missing Data} & \textbf{Key Parameters} \\
 & \textbf{Heads} & \textbf{Loss} & \textbf{Capable} & \textbf{Handling} & \\
\midrule
Baseline NN & Yes & Yes & Yes & Masked loss & Hidden: 128, 64 units \\
             &     &     &     & functions   &  Dropout: 0.2 \\
\addlinespace
Pre-trained MLP & Yes & Yes & Yes & Masked loss & Latent: 128 dims \\
(Autoencoder)     &     &     &     & functions   & Epochs: 100 (pre-train) \\
\addlinespace
FT-Transformer & Yes & Yes & Yes & Masked loss & Blocks: 2, Dim: 192 \\
               &     &     &     & functions   & Epochs: 60 \\
\addlinespace
TabNet & No & No & No & Remove missing & Separate model \\
       &    &    &    & samples        & per toxin \\
\addlinespace
TabPFN & No & No & No & Remove missing & Pre-trained prior \\
       &    &    &    & samples        & No retraining \\
\bottomrule
\end{tabular}
\end{table}

\subsection{Baseline Feedforward Neural Network}
\label{subsec:baseline_fnn}

The baseline model was a fully connected feedforward neural network designed to jointly predict continuous mycotoxin concentrations and their corresponding binary presence–absence indicators. The architecture consisted of a shared backbone (input and shared hidden layers) and two task-specific output heads.

The input layer accepted a one-dimensional feature vector of 454 predictors. This was followed by two fully connected hidden layers with 128 and 64 units, respectively, each using the Rectified Linear Unit (ReLU) activation function \citep{nair2010rectified}. Dropout layers with a rate of 0.2 were applied after each hidden layer to reduce overfitting \citep{srivastava2014dropout}. From the shared backbone, the network branched into two output heads: a continuous head predicting the 24 mycotoxin concentrations using a dense layer with 24 units and ReLU activation, and a binary head predicting the presence or absence of the same 24 toxins using a dense layer with 24 units and sigmoid activation.

As not all toxins were measured for every sample, the response matrices contained missing entries. To avoid discarding entire samples with incomplete responses, custom masked loss functions were implemented. These computed the error only over observed (non-missing) responses for each sample. Missing values were encoded and masked dynamically during training. For the continuous head, a masked Mean Squared Error (MSE) was used:
\begin{equation}
\label{eq:masked_mse}
\mathcal{L}_{\text{MSE}} = \frac{\sum_{i=1}^K m_i (y_i - \hat{y}_i)^2}{\sum_{i=1}^K m_i + \epsilon}
\end{equation}
and for the binary head, a masked Binary Cross-Entropy (BCE):
\begin{equation}
\label{eq:masked_bce}
\mathcal{L}_{\text{BCE}} = - \frac{\sum_{i=1}^K m_i \left[ y_i \log(\hat{y}_i) + (1-y_i) \log(1 - \hat{y}_i) \right]}{\sum_{i=1}^K m_i + \epsilon}
\end{equation}
where $\epsilon = 10^{-7}$ ensured numerical stability, and $\hat{y}_i$ in \eqref{eq:masked_bce} was clipped to $[\epsilon, 1-\epsilon]$ to prevent undefined logarithms. Here, $m_i$ denotes a binary mask indicating whether response value $y_i$ is observed ($m_i = 1$) or not ($m_i = 0$).

The model was trained with the Adam optimiser (learning rate $10^{-3}$), using masked MSE for the continuous outputs and masked BCE for the binary outputs; the same masked losses were tracked as metrics. The dataset was split 80\%–20\% into training and test sets, with 20\% of the training data held out for validation. Training proceeded for up to 500 epochs with a mini-batch size of 32, early stopping on validation loss (patience 25) and restoration of the best weights. To preserve the temporal/block structure in the data, batch shuffling was disabled during training.

\subsection{Pre-trained Feedforward Neural Network (Autoencoder initialisation)}
\label{subsec:transfer_learning_nn}
%\subsection{Transfer Learning Neural Network with Autoencoder Pretraining}
%\label{subsec:transfer_learning_nn}

This model extends the baseline MLP by using autoencoder-based pre-training as transfer learning. An autoencoder (AE) is first trained on the predictor matrix to learn compressed feature representations \citep{hinton2006reducing}; the encoder is then reused to initialise the supervised MLP, which is fine-tuned for the joint regression–classification tasks.

\subsubsection{Autoencoder Pre-training}
An AE with symmetric encoder-decoder architecture was first trained on the entire predictor matrix ($n \times 454$) using a reconstruction objective \citep{lecun2015deep}. The encoder compressed the input features through successive dense layers with 512, 256, and 128 units respectively, each with ReLU activation and dropout (rate 0.2), creating a 128-dimensional latent representation. The decoder symmetrically reconstructed the original input by expanding back up through the layers. This unsupervised pre-training forced the encoder to learn compressed representations that capture the underlying structure of the predictor space \citep{vincent2008extracting}. The AE was trained for up to 100 epochs using MSE loss and early stopping (patience 10).

\subsubsection{Supervised fine-tuning}
Following pre-training, the decoder was discarded and the trained encoder served as a feature extraction backbone for the downstream mycotoxin prediction tasks. Two task specific heads were attached to the encoder's 128-dimensional output. A regression head with ReLU activation predicted continuous mycotoxin concentrations while a classification head with sigmoid activation predicted binary presence-absence. The model employed the same custom masked loss functions described in Section \ref{subsec:baseline_fnn} (that is, Equations~\ref{eq:masked_mse}–\ref{eq:masked_bce}) to handle missing response values.

Two fine-tuning strategies were evaluated. The first approach froze the encoder weights so only the prediction heads were trained, treating the encoder as a static feature extractor. The second approach trained the entire model end-to-end, allowing the encoder representations to adapt specifically for mycotoxin prediction. For more on the performance of freezing layers versus fine-tuning the entire network, see \cite{yosinski2014transferable}.

\subsection{TabNet}
\label{subsec:tabnet}

TabNet is a deep learning architecture specifically designed for learning from tabular datasets, combining the flexibility of neural networks with specialised attention mechanisms for feature selection and interpretability \citep{arik2021tabnet}. TabNet's sequential attention allows it to focus on the most relevant features at each decision step, providing both high predictive performance and insights into feature importance \citep{shwartz2022tabular}. The architecture uses a series of decision steps, each equipped with a feature selection module that learns which columns should be attended to for the current prediction task. The model processes input data sequentially, allowing it to dynamically adapt which features are used for each sample. This design enables TabNet to capture complex, non-linear relationships in tabular data and generalise well even when the number of training samples is limited.

For this study, separate TabNet models were trained for each mycotoxin due to the architecture's single output design. Each toxin required a dedicated TabNet model trained on the pre-split training and test datasets, utilising only numeric features. For binary outcomes, the probabilistic classifier from the \texttt{pytorch-tabnet} library \citep{pytorch_tabnet} was used, while continuous mycotoxin concentrations were modelled with its regressor. Missing response values were filtered out prior to model fitting and evaluation to meet TabNet's requirement for complete response observations. Predictions were generated for the complete test set, with performance metrics computed only on samples with observed response values. The current TabNet implementation does not permit custom output heads or user defined loss functions, consequently all experiments used TabNet's default objectives of cross-entropy loss for classification and mean squared error for regression.

\subsection{FT-Transformer}
\label{subsec:ft_transformer}

The FT-Transformer (Feature Tokeniser Transformer) is a deep learning model that adapts the transformer architecture for tabular data \citep{shwartz2022tabular}. It treats all features as embeddings and processes them through a stack of transformer layers, enabling the model to learn complex, high-order interactions between features. The model begins by converting each feature value into a vector embedding through learned linear projections for continuous features. These feature embeddings are then fed into a series of transformer blocks, which use self-attention mechanisms to weigh and combine information across the entire feature set. 

A key advantage of the FT-Transformer is its architectural flexibility. Unlike models with fixed prediction outputs, its backbone can serve as a universal feature extractor for multiple downstream tasks \citep{shwartz2022tabular}. In this study, we used this capability to build a multi-task learning model. A single FT-Transformer backbone was trained to learn a shared representation of the input features. Two separate prediction heads were then attached to this backbone: a regression head consisting of a linear layer with ReLU activation to predict continuous mycotoxin concentrations for all toxins simultaneously, and a multi-label classification head comprising a linear layer followed by sigmoid activation to predict binary presence-absence labels for all toxins.

The entire network was trained jointly by optimising a loss function combining masked mean squared error for the regression task and masked binary cross-entropy for the classification task, using the \newline \texttt{rtdl\_revisiting\_models} library \cite{rtdl_revisiting_models}. The masked loss functions were identical to those used in the baseline neural network (Equations \ref{eq:masked_mse} and \ref{eq:masked_bce}), enabling the model to learn from all available data while handling samples where some toxin labels were missing. The model was trained for up to 60 epochs with early stopping (patience 8) based on validation loss, using the AdamW optimiser \citep{loshchilov2017decoupled} with a learning rate of $3 \times 10^{-3}$.

\subsection{TabPFN}
\label{subsec:tabpfn}

TabPFN (Tabular Prior-Data Fitted Network) is a NN model designed specifically for tabular data, employing a form of TL tailored for small datasets. Unlike traditional tabular models, TabPFN is trained as a \textit{prior} over a massive number of synthetic tabular classification and regression problems, effectively learning to generalise across diverse tasks with minimal fine-tuning \citep{hollmann2022tabpfn}. At inference time, it can be applied directly to new datasets, providing high quality predictions with limited or no retraining. TabPFN uses a transformer architecture that encodes both the features and labels of the training data, enabling it to learn from millions of simulated datasets. When presented with a new tabular prediction task, TabPFN applies its pre-trained weights to infer the relationship between features and responses.

We applied TabPFN, using the \texttt{tabpfn} library \citep{tabpfn_lib}, to both binary classification and continuous regression endpoints across all mycotoxins. For each toxin, a separate TabPFN model was trained on pre-split training data and evaluated on test data. In the classification setting, probabilistic outputs were generated for each toxin specific binary label. For regression, TabPFN produced continuous value predictions corresponding to measured toxin concentrations. A key constraint of the current TabPFN implementation is the lack of support for custom output heads or user defined loss functions. As a result, all experiments were conducted using the model's default prediction heads and loss formulations for both classification and regression tasks. Consequently, any samples with missing response values were removed prior to model fitting and evaluation, as the standard loss functions cannot handle missing response values.

%%%%%%%%%%%%%%%%

\subsection{Evaluation}
 Continuous outputs were evaluated by root mean squared error (RMSE) and coefficient of determination ($R^2$), while binary outputs were evaluated by F1-score and area under the receiver operating characteristic curve (AUC). Per-toxin metrics were computed and averaged to assess overall performance.

\section{Results}
\label{sec:results}

\subsection{Overall Model Performance}
\label{subsec:overall_performance}

We evaluated model performance using multiple metrics appropriate for each prediction task. For continuous mycotoxin concentrations, we used RMSE and $R^2$. For binary presence or absence predictions, we used F1-score and Area Under the ROC Curve (AUC). Table \ref{tab:model_performance} presents the average performance across all 24 mycotoxins for each model architecture.

The baseline neural network achieved the highest $R^2$ (0.57) among all models and competitive regression accuracy (RMSE = 0.88 \textmu g/kg), ranking closely behind TabPFN. It also maintained strong binary prediction performance with an F1-score of 0.68 and AUC of 0.94. TabPFN achieved the lowest RMSE (0.81 \textmu g/kg) and a comparable $R^2$ (0.56), while outperforming all other models on the classification tasks with the highest F1-score (0.77) and AUC (0.96). Overall, TabPFN delivered the best performance across three of the four evaluation metrics, though the baseline neural network remained close in both regression and classification accuracy. The transfer learning model with unfrozen fine-tuning achieved moderate performance ($R^2$~=~0.47; F1~=~0.70), whereas its frozen variant performed notably worse ($R^2$~=~0.26; F1~=~0.56), reinforcing the importance of end-to-end fine-tuning for effective feature adaptation. The FT-Transformer achieved a reasonable AUC (0.91) but weaker regression and classification accuracy overall. TabNet again underperformed across all metrics, with a large RMSE (8.18) and negative $R^2$ values, suggesting instability under the limited sample size and missing data structure of the mycotoxin dataset.

\begin{table}[H]
\centering
\caption{Average predictive performance across all mycotoxins. Best performance in each metric is highlighted in bold text.}
\label{tab:model_performance}
\begin{tabular}{l|r|r|r|r}
\hline
\textbf{Model} & \textbf{RMSE} & \textbf{$R^2$} & \textbf{F1} & \textbf{AUC} \\
\hline
Baseline Neural Network & 0.88 & \textbf{0.57} & 0.68 & 0.94 \\
TabPFN & \textbf{0.81} & 0.56 & \textbf{0.77} & \textbf{0.96} \\
Transfer Learning (Unfrozen) & 0.92 & 0.47 & 0.70 & 0.94 \\
FT-Transformer & 2.75 & 0.31 & 0.67 & 0.91 \\
Transfer Learning (Frozen) & 1.56 & 0.26 & 0.56 & 0.89 \\
TabNet & 8.18 & -1.79 & 0.30 & 0.58 \\
\hline
\end{tabular}
\end{table}

\subsection{Per-Toxin Performance}
\label{subsec:per_toxin}

\begin{figure}[H]
\centering
\includegraphics[width=0.8\textwidth]{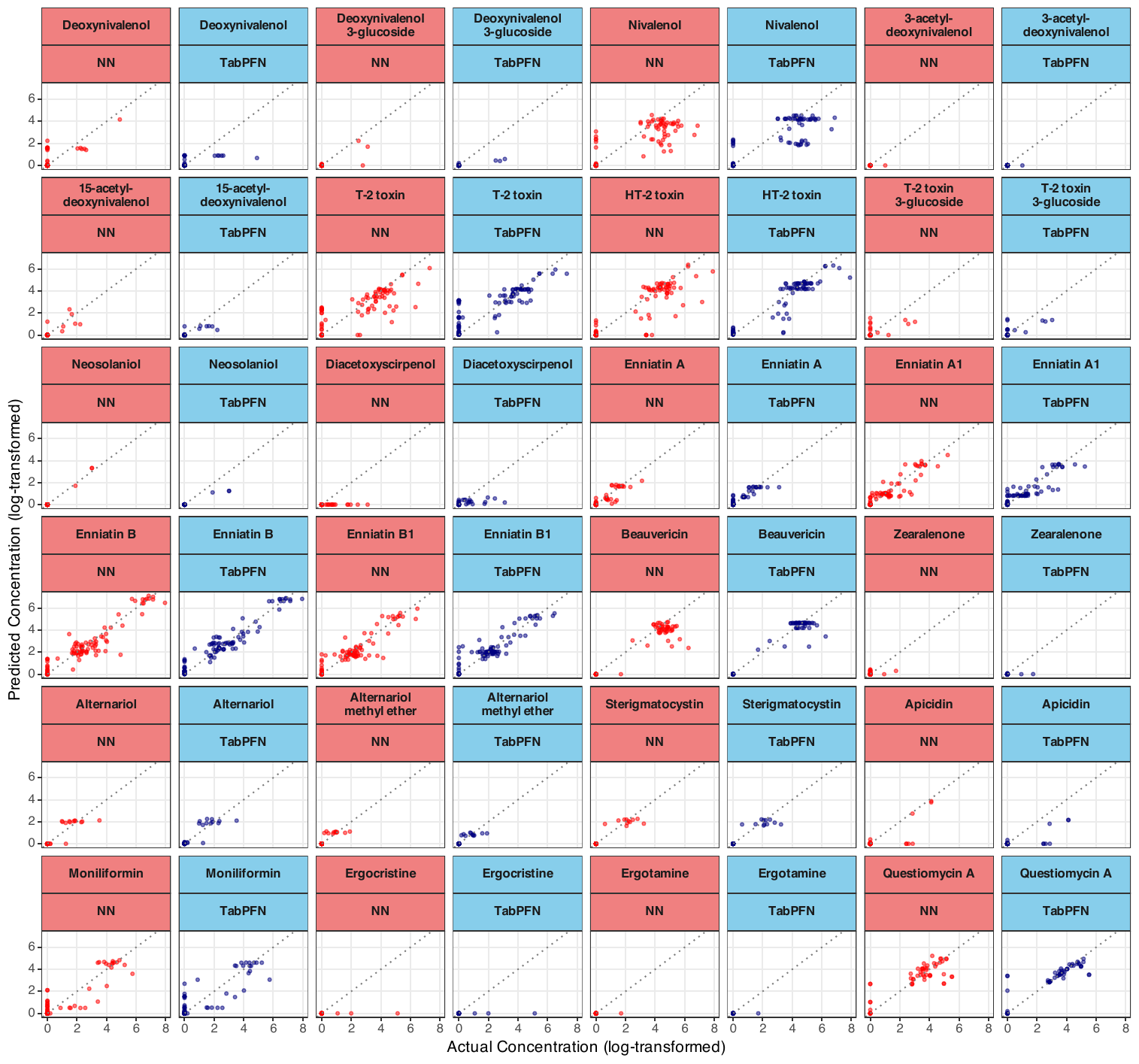}
\caption{Actual versus predicted values for each continuous mycotoxin from the baseline NN (in red) and TabPFN (in blue) models.}
\label{fig:avp_cont}
\end{figure}

\begin{figure}[H]
\centering
\includegraphics[width=0.8\textwidth]{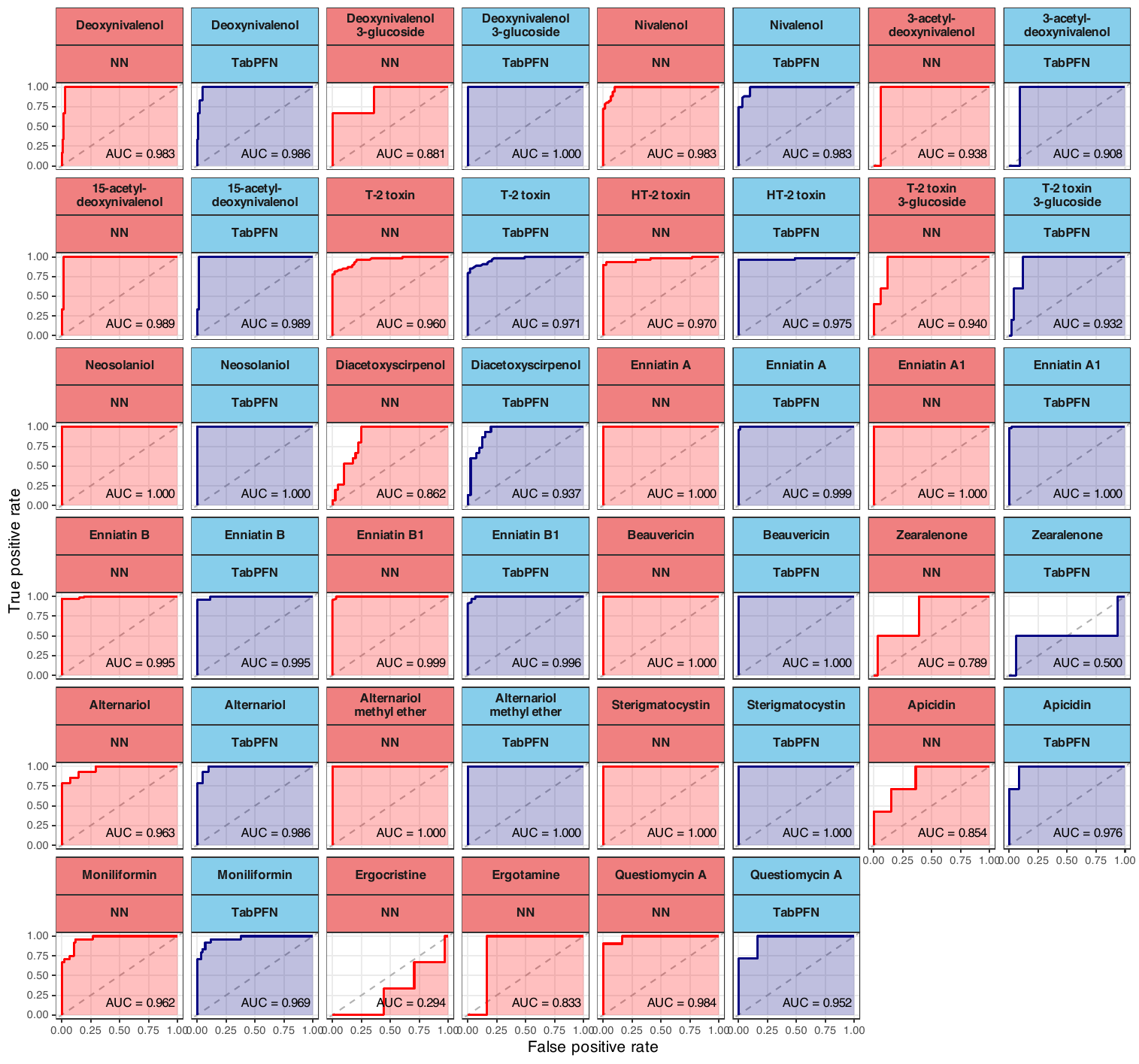}
\caption{Receiver operating characteristic (ROC) curves for binary mycotoxin predictions on the test set for both the baseline NN (in red) and the TabPFN (in blue) models.}
\label{fig:avp_bin}
\end{figure}

While the overall performance metrics provide a general comparison, individual mycotoxins showed substantial variation in predictability across models. Figure~\ref{fig:avp_cont} shows actual versus predicted values for continuous concentrations using the baseline neural network and the TabPFN models on the test data, while Figure~\ref{fig:avp_bin} displays Receiver Operating Characteristic (ROC) curves for both the baseline neural network and the TabPFN model. These two models are highlighted as they achieved the highest overall performance, with the neural network giving the best average $R^2$ across toxins and TabPFN yielding the best average AUC.

TabPFN delivered very strong discrimination for many toxins (AUC often~$\geq 0.97$; for example ENN~A, ENN~A1, ENN~B, ENN~B1, beauvericin, HT-2~toxin, and alternariol), and frequently matched or exceeded the baseline. For regression, the enniatins and beauvericin were among the most predictable (for example, TabPFN achieved $R^2$ values of 0.83 for ENN~A, 0.82 for ENN~A1, 0.92 for ENN~B, 0.86 for ENN~B1, and 0.93 for beauvericin), whereas several deoxynivalenol-related toxins and zearalenone showed weaker fits and tended to favour the baseline model. For instance, 15-acetyl-deoxynivalenol and deoxynivalenol-3-glucoside had lower $R^2$ values from TabPFN (0.55 and 0.29) compared with the baseline (0.65 and 0.57, respectively). In contrast, TabPFN showed clear AUC gains for some toxins where the baseline was modest (for example, AUC increased from 0.88 to 1.00 for deoxynivalenol-3-glucoside and from 0.85 to 0.98 for apicidin). Two toxins (ergocristine and ergotamine) lacked valid binary results for TabPFN due to having only one outcome class in the test set (for example, all positives or all negatives), so they are omitted from Figure~\ref{fig:avp_bin}. As previously mentioned, a few high scores occur for toxins with limited test samples, which can inflate both~$R^2$ and AUC, as such these cases should be interpreted cautiously.

\begin{figure}[H]
\centering
\includegraphics[width=0.49\textwidth]{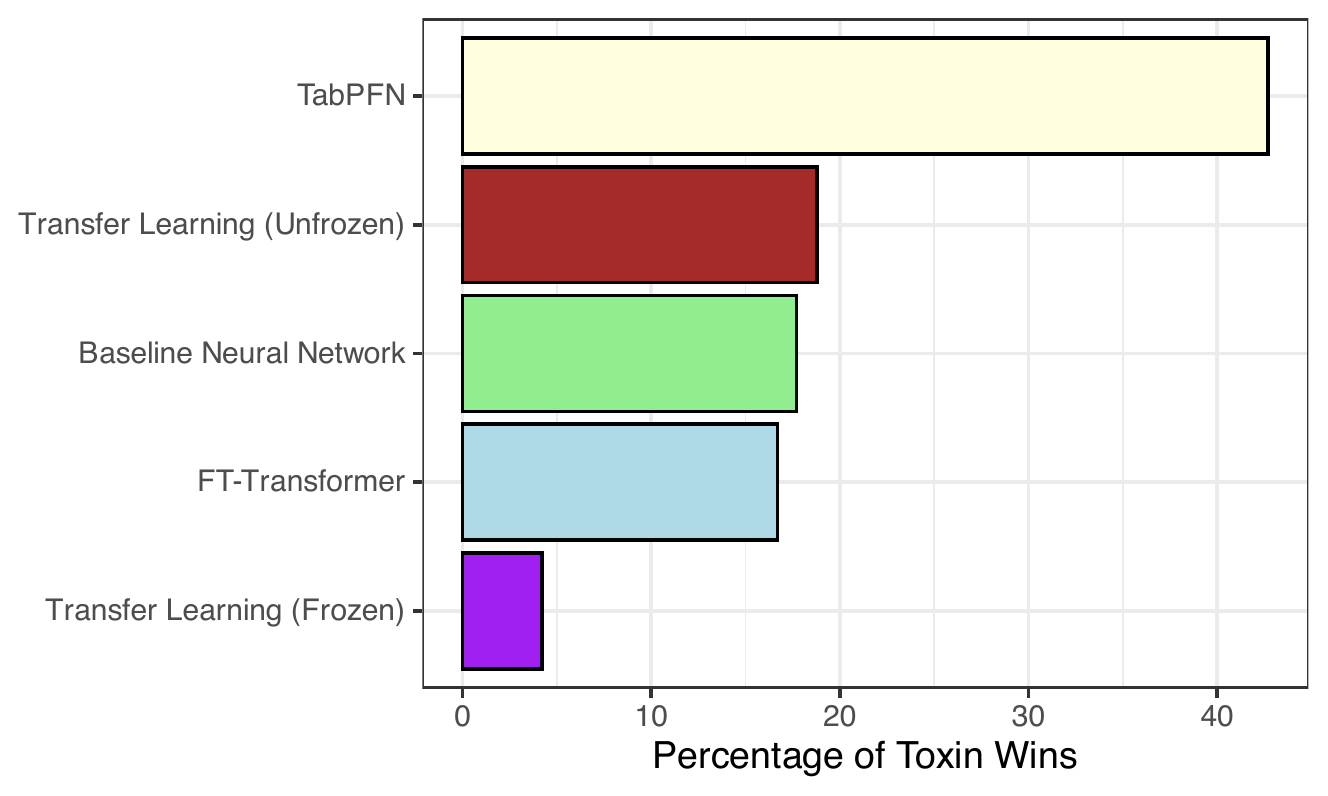}
\caption{Overall model performance ranking by percentage of total wins across all toxin-metric combinations. TabPFN achieves 42.7\% of all wins, demonstrating overall superiority while highlighting competitive performance from Transfer Learning and traditional neural network approaches.}
\label{fig:overall_winners}
\end{figure}

Although the Baseline NN shows the best average $R^2$ performance, patterns vary when examining individual toxin–metric combinations. We ranked models using a winner-takes-all approach where the optimal performing model (lowest RMSE or highest $R^2$/F1/AUC) for each toxin–metric pair was designated the \textit{winner}, regardless of absolute performance level. Figure~\ref{fig:overall_winners} provides the aggregate ranking across all toxin–metric combinations, with TabPFN achieving 42.7\% of total wins, followed by TL~Unfrozen~(18.8\%), Baseline~NN~(17.7\%), FT–Transformer~(16.7\%), and TL~Frozen~(4.2\%). TabNet is not shown, as it did not achieve the highest score for any toxin–metric pair.

\subsection{Variable Importance}
\label{subsec:vimp}

To assess which predictor variables had the most influence on the responses, a model-agnostic variable importance (Vimp) method was used. Specifically, a permutation-based Vimp \citep{breiman2001random} was applied. In this approach, the values of a given predictor are randomly shuffled across samples, thereby breaking its association with the response while preserving the overall distribution of the variable. The model is then re-evaluated on the permuted data, and the drop in predictive performance is recorded. A larger decrease in performance indicates that the variable contributed more strongly to the model's predictions \citep{strobl2008conditional, fisher2019all}. This procedure is repeated across variables, providing a relative ranking of importance that is independent of model architecture. For more information on variable importance methods in general see \cite{ inglis2022visualizing, molnar2020interpretable}. The permutation importance calculations were implemented using the same custom masked loss functions used during model training and evaluation. This ensured consistent treatment of missing values throughout the analysis, with importance scores reflecting the actual performance metrics optimised during model development (masked MSE for regression and masked BCE for classification tasks).

\begin{figure}[H]
\centering
\includegraphics[width=0.8\textwidth]{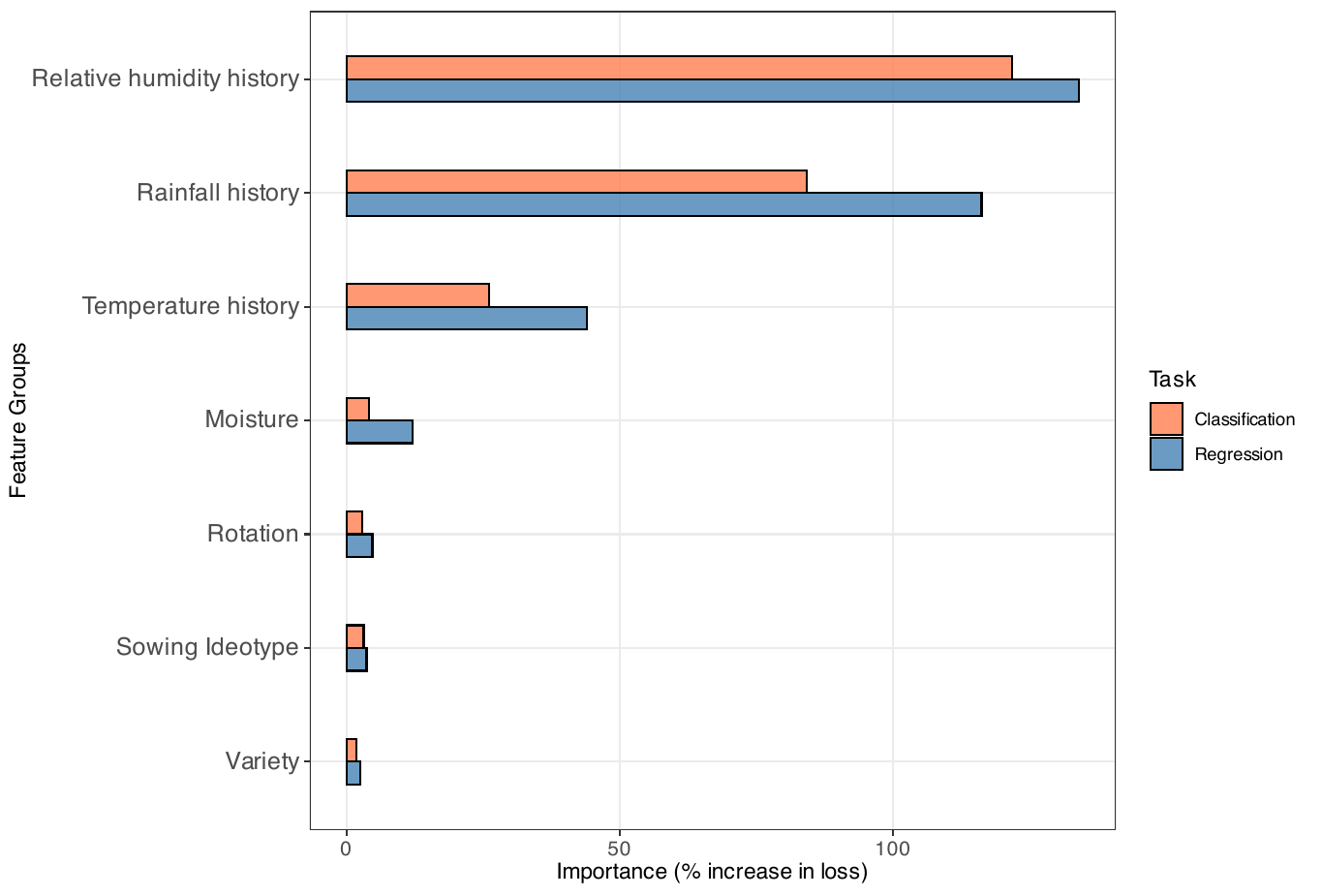}
\caption{Grouped permutation variable importance rankings for regression and classification tasks. Importance scores are based on masked Mean Squared Error (MSE) for regression and masked Binary Cross Entropy (BCE) for classification. Weather history variables (humidity, rainfall, temperature) consistently rank as the most influential predictors across both modelling objectives.}
\label{fig:vimp_grouped}
\end{figure}

Given that one-hot encoded categorical variables can produce misleading importance scores when permuted individually (as this breaks the categorical structure), a grouped permutation approach was implemented. This method permutes all dummy variables belonging to the same original categorical variable simultaneously, preserving the categorical relationships and providing more meaningful importance assessments for multi-level factors such as geographic counties and management practices. Figure \ref{fig:vimp_grouped} displays the variables that ranked in the top ten for both regression and classification analyses using this grouped approach, of which seven were common across tasks. Weather history variables dominate these shared rankings, with humidity, rainfall, and temperature history emerging as the three most influential predictors in both loss types, underscoring the critical role of weather patterns during the 90-day pre-harvest period. Seed moisture also appears as an important predictor, followed by rotation, sowing ideotype, and variety.

\begin{figure}[H]
\centering
\begin{minipage}{0.68\textwidth}
  \centering
  \includegraphics[width=\textwidth]{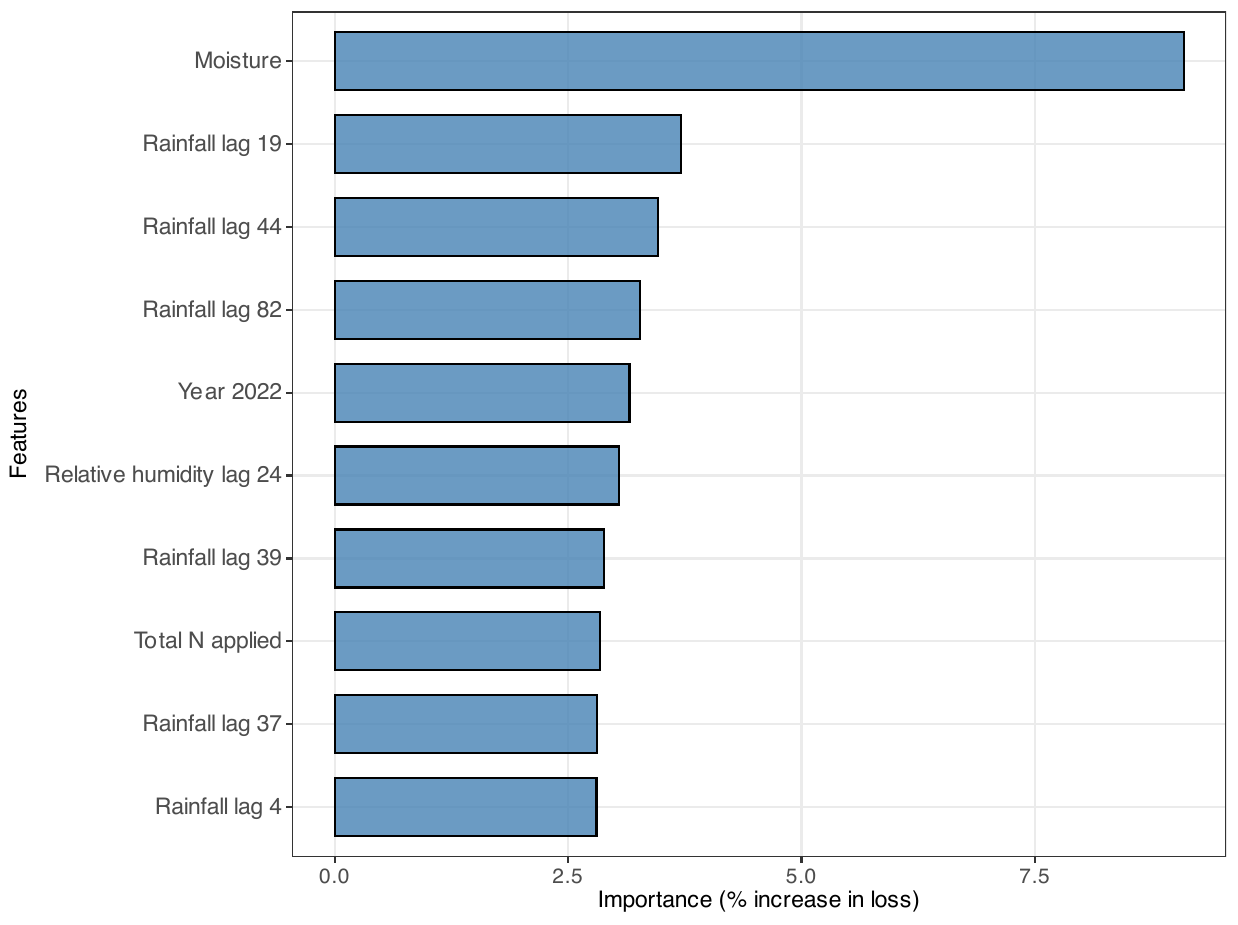}
\end{minipage}
\hfill
\begin{minipage}{0.68\textwidth}
  \centering
  \includegraphics[width=\textwidth]{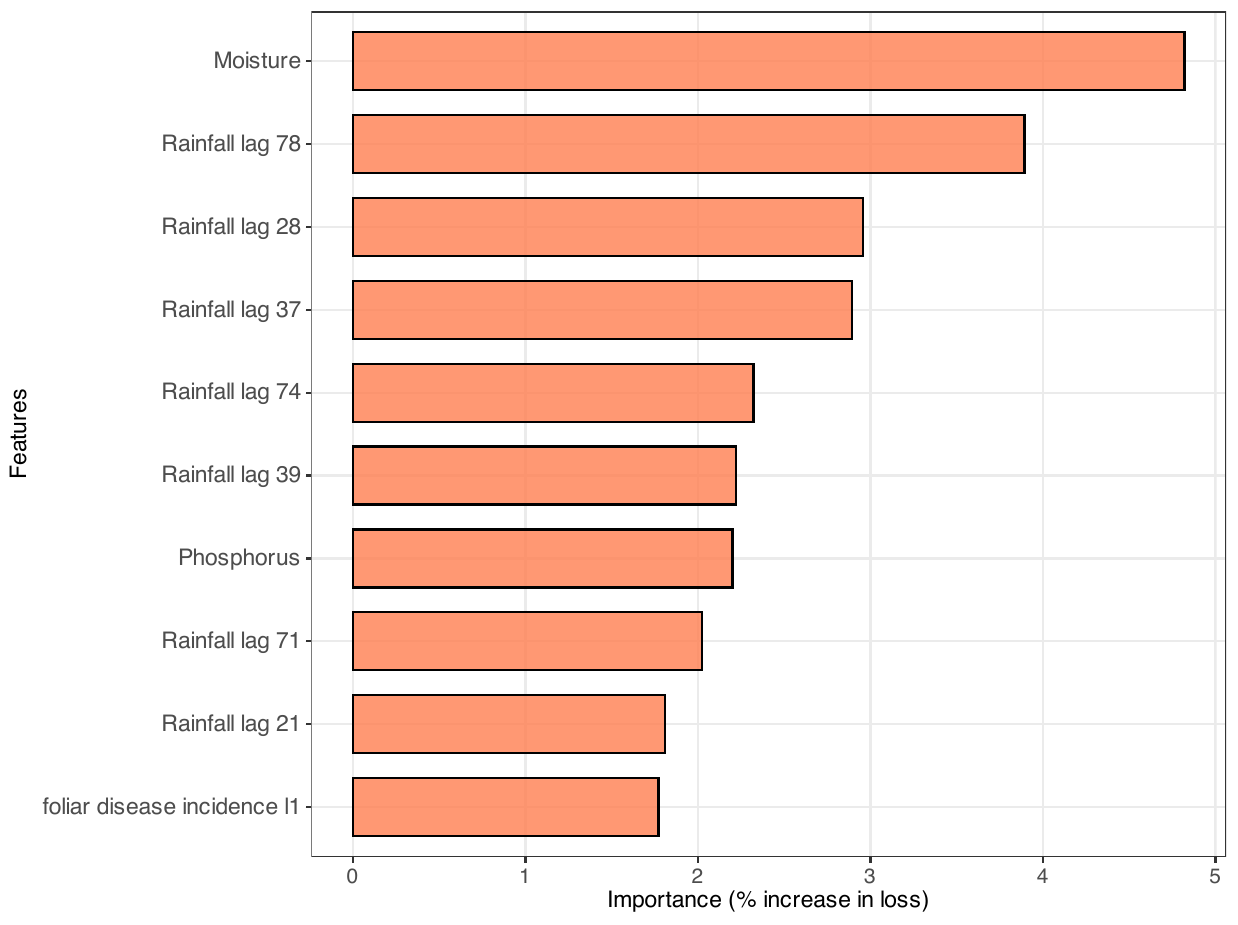}
\end{minipage}
\caption{Feature importance for the top 10 predictors in the regression (blue) and classification (red) tasks. Weather-related variables dominate both, with seed moisture showing consistently high importance.}
\label{fig:vimp_ind}
\end{figure}

To gain further granularity, specifically regarding levels within categorical variables, importance was also assessed for individual one-hot encoded features. Figure \ref{fig:vimp_ind} shows the importance for these individual features that ranked highly in both regression (top plot in blue) and classification tasks (bottom plot in red). This analysis treats each dummy variable from categorical factors as a separate feature. Seed moisture content emerged as the single most important predictor, showing a 9.1\% increase in regression loss and 4.8\% increase in classification loss when permuted. For both regression and classification loss, rainfall timing variables dominated the top 10 most important variables. The consistency of these rankings across both tasks suggests these variables have fundamental importance for mycotoxin prediction regardless of the modelling objective.

\subsection{Method Performance (Validation)}

Full validation results, including linearity, recovery, and precision metrics for all analytes, are provided in the Supplementary Materials. The within-laboratory reproducibility ($RSD_{WLR}$) met the performance criterion ($\le 20\%$) for the majority of mycotoxins; however, ergocristine and nivalenol failed to meet this standard. Regarding extraction efficiency, most analytes demonstrated recovery rates within the ranges mandated by EC decision 2002/657 \citep{eurpean2002commission} (70--120\% for quantitative methods, with 60--140\% permitted for routine analysis). Ergocristine fell outside these limits, a result attributed to the sub-optimal pH of the extraction solvent for this specific alkaloid. Consequently, results for ergocristine and nivalenol are considered qualitative and suitable for screening purposes only. The validation data confirm that the method is fit-for-purpose for the quantitative analysis of the remaining target mycotoxins.

%The analytical method was validated for linearity, recovery, and precision in accordance with SANTE/11312/2021 guidelines. Full validation results, including specific metrics for all analytes, are provided in the Supplementary Materials. Apparent recovery and within-laboratory reproducibility ($RSD_{WLR}$) were calculated based on 75 samples (five oat lots spiked in quintuplicate across three days). The repeatability ($RSD_{WLR}$) met the performance criterion ($\le 20\%$) for the majority of mycotoxins, with the exception of ergocristine and nivalenol. The efficiency of the extraction process was assessed by comparing spiked oat samples against spiked blank extracts ($n=25$ each). Recovery rates were evaluated against the criteria set out in EC decision 2002/657 \citep{eurpean2002commission}, which requires a range of 70--120\% for quantitative methods (with 60--140\% permitted for routine analysis), with most analytes within this range. However, ergocristine fell outside these limits. For ergocristine, this reduced recovery was attributed to the pH of the extraction solvent being sub-optimal for these specific compounds. Consequently, results for this ergot alkaloid, and the mycotoxin nivalenol, should be considered qualitative and suitable for screening purposes only. Overall, the validation data confirm that the method is fit-for-purpose for the quantitative analysis of the remaining target mycotoxins.

\section{Conclusion}
\label{sec:conclusion}

Our study evaluated the application of NN and TL approaches for predicting mycotoxin contamination in Irish oats, comparing five distinct modelling architectures across both regression and classification tasks. The research addresses a critical agricultural challenge where accurate mycotoxin prediction can support early intervention strategies and reduce economic losses in cereal production systems. The transfer learning model, TabPFN, achieved the strongest overall performance, outperforming the other architectures across three of the four evaluation metrics (RMSE, F1, and AUC). While the baseline feedforward NN remained highly competitive, achieving the highest $R^2$, the superior results of TabPFN were further reinforced by the individual toxin-metric comparisons, where it secured 42.7\% of total wins. This suggests that pre-trained models using knowledge from diverse tabular datasets can effectively generalise to mycotoxin prediction problems, even with limited training data. Furthermore, for the custom neural network architectures, the implementation of masked loss functions proved essential for handling missing value patterns, allowing these models to learn from all available information without discarding incomplete samples.

Variable importance analysis revealed that weather history patterns in the 90-day pre-harvest period were the most important predictors of mycotoxin contamination in our dataset (when grouping dummy variables belonging to the same original categorical variable). Humidity history was the most important for both regression and classification tasks, followed by rainfall and temperature history. At the level of individual predictors, seed moisture content emerged as the single most important variable for both tasks, followed by specific rainfall timings. These model-derived findings are based on a limited sample size and should be interpreted as indicative rather than exhaustive. Nonetheless, they are consistent with established knowledge of the environmental conditions that favour fungal growth and mycotoxin production \citep{platzer2025investigating, leggieri2020impact, nazari2014influence, paterson2010will}.

While TL approaches showed promise, particularly TabPFN's strong classification performance, the results also highlight important limitations. TabNet performed poorly across all metrics, likely reflecting sensitivity to small sample sizes and missing data. The FT-Transformer showed moderate performance, suggesting that while transformer architectures can be adapted for mycotoxin prediction, they may require further optimisation for this specific domain. Geographic specificity also presents limitations for generalisation to other regions with different climatic conditions and agricultural practices. Future research should explore model transferability across diverse geographic and climatic contexts, potentially using the TL approaches evaluated here.

Several avenues for future work emerge from this research. First, the incorporation of additional data sources, such as remote sensing imagery, soil microbiome profiles, and real-time sensor data, could enhance predictive accuracy. Second, the development of ensemble approaches combining the strengths of different architectures might yield superior performance to individual models. Finally, the integration of these predictive models into decision support systems for farmers and food safety authorities represents a critical step toward practical implementation. Such a system would need to account for the dynamic nature of key predictors. In a prospective setting, weather-related variables would require continuous updates to reflect current conditions, whereas many agronomic and geographical variables would remain constant. Although this study is Ireland-focused, the modelling framework and transfer learning approaches demonstrated here could be adapted to support mycotoxin risk prediction in other regions, particularly in resource-limited settings where laboratory capacity and surveillance infrastructure are constrained.

\section*{Funding}
This work was conducted as part of the Mycotox-I project, which is kindly supported by the Department of Agriculture, Food, and the Marine (DAFM) and the Department of Agriculture, Environment, and Rural Affairs (DAERA), grant number 2021/R/460.

\bibliographystyle{unsrtnat}              
\bibliography{ref}

\end{document}

\begin{center}
{\large\bf SUPPLEMENTAL MATERIALS}
\end{center}

The base NN and keras TL (both frozen and unfrozen) were made using \texttt{R}, the remaining TL models were all created using \texttt{Python}. Although the data for this project is proprietary and can not be publicly shared, the full coding scripts for the models can be found here: \url{https://github.com/AlanInglis/Mycotox-I}

\subsection*{S1: Mycotoxins Used In Machine Learning Models}

\setcounter{table}{0}% start from 0 
\renewcommand{\thetable}{S1.\arabic{table}}%

\begin{table}[H]
\centering
\begin{tabular}{l|l}
\hline
\textbf{Mycotoxin} & \textbf{Mycotoxin} \\
\hline
Deoxynivalenol & Enniatin A \\
Deoxynivalenol-3-glucoside & Enniatin A1 \\
Nivalenol & Enniatin B \\
3-acetyl-deoxynivalenol & Enniatin B1 \\
15-acetyl-deoxynivalenol & Beauvericin \\
T-2 toxin & Zearalenone \\
HT-2 toxin & Apicidin \\
T-2 toxin-3-glucoside & Sterigmatocystin \\
Neosolaniol & Diacetoxyscirpenol \\
Questiomycin A & Alternariol \\
Alternariol methyl ether & Moniliformin \\
Ergocristine & Ergotamine \\
\hline
\end{tabular}
\caption{List of mycotoxins measured in this work.}
\label{tab:mycotoxin_list}
\end{table}

\begin{table}[H]
\centering
\caption{Summary statistics of mycotoxin concentrations and detection rates used in the machine learning model.}
\label{tab:summary_stats}
\small
\begin{tabular}{p{4.8cm}|l|l|l|l|l|l|l}
\toprule
\textbf{Mycotoxin} & \textbf{Mean} & \textbf{SD} & \textbf{Min} & \textbf{Max} & \textbf{Occurrence} & \textbf{Total} & \textbf{Occurrence (\%)} \\
\midrule
Enniatin B                  & 170.00 & 386.00 & 0.00 & 2852.00 & 423 & 549 & 77.0 \\
Questiomycin A              &  47.20 &  50.40 & 0.00 &  253.00 & 232 & 318 & 73.0 \\
Enniatin B1                 &  48.70 & 126.00 & 0.00 & 1656.00 & 394 & 549 & 71.8 \\
Beauvericin                 &  60.30 &  65.10 & 0.00 &  523.00 & 231 & 390 & 59.2 \\
HT-2 toxin                  &  77.90 & 228.00 & 0.00 & 3127.00 & 295 & 548 & 53.8 \\
Enniatin A1                 &  10.20 &  29.90 & 0.00 &  392.00 & 286 & 548 & 52.2 \\
T-2 toxin                   &  39.50 & 119.00 & 0.00 & 1681.00 & 299 & 581 & 51.5 \\
Nivalenol                   &  39.70 &  78.00 & 0.00 &  967.00 & 236 & 597 & 39.5 \\
Enniatin A                  &   1.38 &   6.33 & 0.00 &  122.00 & 149 & 549 & 27.1 \\
Alternariol                 &   3.05 &  11.00 & 0.00 &  112.00 &  90 & 342 & 26.3 \\
Moniliformin                &  18.40 &  46.60 & 0.00 &  369.00 & 155 & 597 & 26.0 \\
Sterigmatocystin            &   2.54 &   7.93 & 0.00 &   69.70 &  80 & 319 & 25.1 \\
Diacetoxyscirpenol          &   0.39 &   1.72 & 0.00 &   21.40 &  69 & 342 & 20.2 \\
Alternariol methyl ether    &   0.57 &   2.58 & 0.00 &   38.60 &  74 & 501 & 14.8 \\
T-2 toxin-3-glucoside       &   1.65 &   7.25 & 0.00 &   56.50 &  47 & 342 & 13.7 \\
Apicidin                    &   3.65 &  15.90 & 0.00 &  153.00 &  32 & 318 & 10.1 \\
Deoxynivalenol              &   2.27 &  11.80 & 0.00 &  133.00 &  64 & 693 &  9.2 \\
Neosolaniol                 &   0.82 &   3.71 & 0.00 &   23.20 &  19 & 318 &  6.0 \\
15-acetyl-deoxynivalenol    &   0.30 &   1.35 & 0.00 &    8.30 &  23 & 390 &  5.9 \\
Deoxynivalenol-3-glucoside  &   1.05 &   6.27 & 0.00 &  101.00 &  24 & 501 &  4.8 \\
Zearalenone                 &   0.72 &   7.03 & 0.00 &  132.00 &  17 & 390 &  4.4 \\
3-acetyl-deoxynivalenol     &   0.12 &   0.71 & 0.00 &    7.11 &  13 & 390 &  3.3 \\
Ergocristine                &   0.86 &  11.70 & 0.00 &  169.00 &   4 & 208 &  1.9 \\
Ergotamine                  &   0.02 &   0.31 & 0.00 &    4.48 &   1 & 208 &  0.5 \\
\bottomrule
\end{tabular}
\end{table}

\subsection*{S2: Predictor Variables}

\setcounter{table}{0}% start from 0 
\renewcommand{\thetable}{S2.\arabic{table}}%

\begin{table}[H]
\centering
\footnotesize
\begin{tabular}{l|l|l}
\hline
\textbf{Variable} & \textbf{Description} & \textbf{Units} \\
\hline
Latitude & Geographic latitude & decimal degrees \\
Longitude & Geographic longitude & decimal degrees \\
Seed Rate & Rate of seed sown & kg/ha \\
Year & Year of the trial & – \\
County & County location of trial & – \\
Sowing Ideotype & Sowing type & Spring or Winter \\
Variety & Crop variety & – \\
Rotation & Previous crop rotation type & – \\
Previous Years (1–5) & Crops grown in the previous five years & – \\
Establishment System & Soil establishment or tillage system & Plough or Min Till \\
Cropping System & Cropping management system & Conventional/Organic \\
Soil Type & Soil texture class & – \\
Soil pH & Soil acidity or alkalinity & pH \\
Phosphorus & Soil phosphorus index & 1–4 \\
Potassium & Soil potassium index & 1–4 \\
Fertiliser Product & Type of fertiliser applied & – \\
Micronutrients Product & Type of micronutrients applied & – \\
Herbicide Product & Type of herbicide applied & – \\
Fungicide Product & Type of fungicide applied & – \\
Growth Regulator Product & Type of growth regulator applied & – \\
Total N Applied & Total nitrogen applied at the beginning of the season & kg/ha \\
Fertiliser Application Rate & Application rate of fertiliser products & kg/ha \\
Micronutrients Application Dose & Application dose of micronutrients & kg/ha \\
Herbicide Dose & Application dose of herbicides & L/ha \\
Fungicide Dose & Application dose of fungicides & L/ha \\
Growth Regulator Dose & Application dose of growth regulators & L/ha \\
Sowing Date & Calendar day of sowing & day of year \\
Fertiliser Application Time & Timing of fertiliser application & day of year \\
Micronutrients Application Date & Timing of micronutrients application & day of year \\
Herbicide Application Time & Timing of herbicide application & day of year \\
Fungicide Application Time & Timing of fungicide application & day of year \\
Growth Regulator Application Time & Timing of growth regulator application & day of year \\
Harvest Date & Calendar day of harvest & day of year \\
Yield & Crop yield at harvest & t/ha \\
Precipitation History & Daily rainfall over the 90 days prior to harvest & mm \\
Temperature History & Daily mean temperature over the 90 days prior to harvest & °C \\
Relative Humidity History & Daily relative humidity over the 90 days prior to harvest & \% \\
\hline
\end{tabular}
\caption{Predictor variables used in the models.}
\label{tab:predictors}
\end{table}

\subsection*{S3: Model Winners per Toxin}
\setcounter{figure}{0}% start from 0 
\renewcommand{\thefigure}{S3.\arabic{figure}}%

\begin{figure}[H]
\centering
\includegraphics[width=0.8\textwidth]{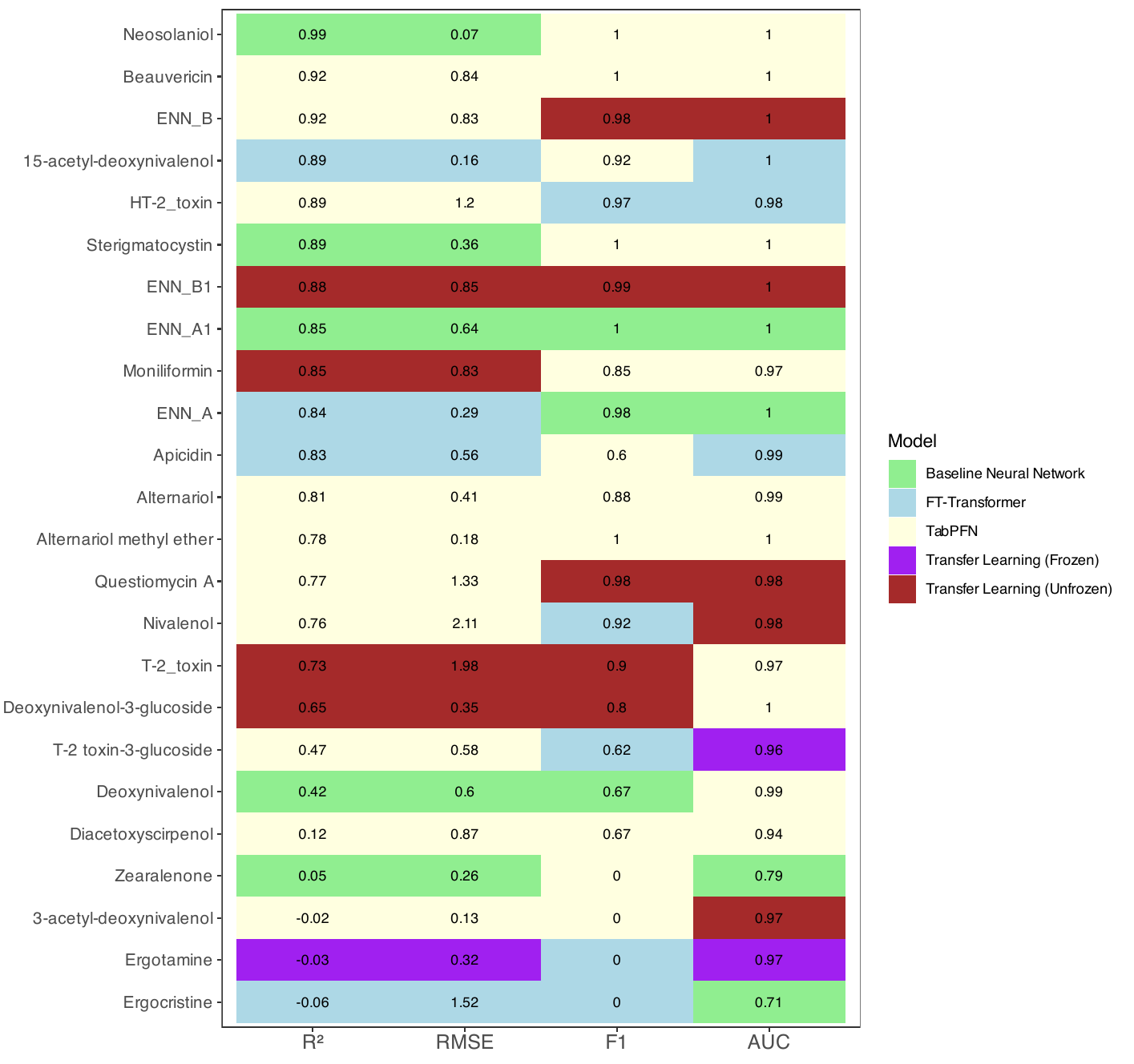}
\caption{Best performing model for each mycotoxin-metric combination. Cell colours indicate the winning model while values show the optimal metric score achieved, sorted by $R^2$ at top-left and descending. Winning models are determined by relative performance comparison (lowest RMSE, highest $R^2$/$F1$/AUC) for each toxin-metric pair, regardless of absolute performance level.}
\label{fig:winners}
\end{figure}

Figure \ref{fig:winners} shows a heatmap of winning models across all combinations, ranked by $R^2$. TabPFN leads in regression and classification, though some wins occur with low absolute performance (e.g., $R^2 < 0.5$), highlighting toxins that are difficult to predict. The Baseline NN follows with 5 regression wins in RMSE and $R^2$, while TL variants remain competitive in classification.

\subsection*{S4: Oat Sample Analysis and Method Validation}

\begingroup
\setcounter{table}{0}% start from 0 so first table is S4.1
\renewcommand{\thetable}{S4.\arabic{table}}%

% Table 1
\begin{longtable}{@{} p{4cm} |p{2.2cm}| p{2.2cm}| p{0.8cm} |p{2.2cm}| r |p{0.8cm}| p{1.1cm} l@{}}
\caption{Details of the optimised MS/MS parameters for the analysed compounds included in the modelling.}\\
\toprule
Analyte & Quantifier Ion (Q) Q1 (m/z) & Qualifier Ion (q) Q3 (m/z) & DP (V) & Collision Energy (V) & CXP (V) & RT (min) & Polarity \\
\midrule
\endfirsthead
\multicolumn{8}{@{}l}{\textit{Table \thetable{} continued}}\\
\toprule
Analyte & Quantifier Ion (Q) Q1 (m/z) & Qualifier Ion (q) Q3 (m/z) & DP (V) & Collision Energy (V) & CXP (V) & RT (min) & Polarity \\
\midrule
\endhead
\bottomrule
\endfoot
15-acetyl-deoxynivalenol & 339.1 $>$ 321.3 & 339.1 $>$ 261.1 & 81 & 13/17 & 18/14 & 4.53 & ESI+ \\
Enniatin A & 699.4 $>$ 682.4 & 699.4 $>$ 210.2 & 11 & 27/39 & 24/22 & 9.64 & ESI+ \\
Enniatin B & 657.3 $>$ 640.3 & 657.3 $>$ 196.1 & 11 & 27/39 & 22/10 & 9.05 & ESI+ \\
Enniatin B1 & 671.3 $>$ 654.4 & 671.3 $>$ 196.1 & 6 & 27/41 & 22/22 & 9.25 & ESI+ \\
Enniatin A1 & 685.4 $>$ 668.5 & 685.4 $>$ 210.0 & 11 & 27/39 & 12/10 & 9.41 & ESI+ \\
Sterigmatocystin & 325.0 $>$ 310.1 & 325.0 $>$ 281.1 & 121 & 35/41 & 16/41 & 7.75 & ESI+ \\
Ergocristine & 610.4 $>$ 592.4 & 610.4 $>$ 223.3 & 76 & 21/47 & 18/12 & 5.2 & ESI+ \\
Beauvericin & 801.3 $>$ 784.3 & 801.3 $>$ 244.1 & 141 & 27/43 & 14/12 & 9.35 & ESI+ \\
HT-2 toxin & 442.3 $>$ 263.1 & 442.3 $>$ 215.1 & 71 & 19/19 & 14/22 & 6.12 & ESI+ \\
T-2 toxin & 484.3 $>$ 215.2 & 484.3 $>$ 185.1 & 76 & 29/31 & 18/11 & 6.65 & ESI+ \\
Diacetoxyscirpenol & 384.2 $>$ 307.3 & 384.2 $>$ 247.3 & 71 & 17/21 & 16/14 & 5.25 & ESI+ \\
Ergotamine & 582.2 $>$ 223.2 & 582.2 $>$ 208.1 & 171 & 43/49 & 10/24 & 4.74 & ESI+ \\
T-2 toxin-3-glucoside & 646.2 $>$ 215.2 & 646.2 $>$ 185.1 & 86 & 35/41 & 12/10 & 6.05 & ESI+ \\
3-acetyl-deoxynivalenol & 397.3 $>$ 59.2 & 397.3 $>$ 307.1 & -60 & -32/-20 & -8/-7 & 4.55 & ESI- \\
Nivalenol & 371.1 $>$ 281.1 & 371.1 $>$ 59.1 & -90 & -20/-50 & -15/-7 & 3.03 & ESI- \\
Alternariol & 256.9 $>$ 213.0 & 256.9 $>$ 215.0 & -125 & -34/-36 & -19/-17 & 6.75 & ESI- \\
Zearalenone & 317.1 $>$ 175.0 & 317.1 $>$ 131.1 & -100 & -34/-42 & -13/-8 & 7.54 & ESI- \\
Moniliformin & 97.0 $>$ 41.2 & n/a & -5 & -38 & -14 & 2.25 & ESI- \\
Alternariol methyl ether & 271.1 $>$ 256.0 & 271.1 $>$ 227.0 & -85 & -32/-50 & -13/-9 & 8.10 & ESI- \\
Deoxynivalenol & 355.1 $>$ 59.0 & 355.1 $>$ 295.1 & -75 & -46/-14 & -7/-21 & 3.65 & ESI- \\
Deoxynivalenol-3-glucoside & 517.2 $>$ 427.1 & 517.2 $>$ 457.2 & -115 & -30/-20 & -11/-19 & 3.55 & ESI- \\
\end{longtable}

\noindent\small\textbf{Notes:} \textsuperscript{13}C-labelled internal standards indicated where shown; n/a: not applicable; Q1: precursor ion, Q3: product ion; CE: collision energy (V); DP: declustering potential (V); RT: retention time; CXP: collision cell exit potential (V); ESI+: Electrospray ionisation positive polarity; ESI-: Electrospray ionisation negative polarity. All sMRM detection windows set at 40 seconds, except for the stable isotopically labelled internal standards which are set at 30 seconds.

\vspace{1em}

% Table 2
\begin{table}[H]
\centering
\caption{Validation spiking levels applied for oats.}
\begin{tabular}{@{}l |r |r@{}}
\toprule
Mycotoxin & Low Level Spike (\textmu g/kg) & High Level Spike (\textmu g/kg) \\
\midrule
15-acetyl-deoxynivalenol & 40 & 200 \\
3-acetyl-deoxynivalenol & 40 & 200 \\
Alternariol & 40 & 200 \\
Alternariol methyl ether & 40 & 200 \\
Beauvericin & 20 & 100 \\
Deoxynivalenol & 40 & 200 \\
Deoxynivalenol-3-glucoside & 20 & 100 \\
Diacetoxyscirpenol & 40 & 200 \\
Enniatin A & 20 & 100 \\
Enniatin A1 & 20 & 100 \\
Enniatin B & 20 & 100 \\
Enniatin B1 & 20 & 100 \\
Ergocristine & 40 & 200 \\
Ergotamine & 40 & 200 \\
HT-2 toxin & 40 & 200 \\
Moniliformin & 40 & 200 \\
Nivalenol & 40 & 200 \\
Sterigmatocystin & 20 & 100 \\
T-2 toxin & 40 & 200 \\
T-2 toxin-3-glucoside & 40 & 200 \\
Zearalenone & 40 & 200 \\
\bottomrule
\end{tabular}
\end{table}

% Table 2
%\begin{table}[H]
%\centering
%\caption{Validation spiking levels applied for oats, barley and poultry feed}
%\begin{tabular}{@{}l |r |r@{}}
%\toprule
%Mycotoxin & HL Spike (\textmu g/kg) & LL Spike (\textmu g/kg) \\
%\midrule
%15-Acetyl-DON & 200 & 40 \\
%3-Acetyl-DON & 200 & 40 \\
%Alternariol (AOH) & 200 & 40 \\
%Alternariol monomethyl ether (AME) & 200 & 40 \\
%Deoxynivalenol (DON) & 200 & 40 \\
%Diacetoxyscirpenol (DAS) & 200 & 40 \\
%Ergocornine & 200 & 40 \\
%Ergocristine & 200 & 40 \\
%Ergotamine (EGT) & 200 & 40 \\
%Fumonisin B1 (FB1) & 200 & 40 \\
%Fumonisin B2 (FB2) & 200 & 40 \\
%HT-2 & 200 & 40 \\
%T-2 toxin & 200 & 40 \\
%T-2-Glucoside (T2G) & 200 & 40 \\
%Moniliformin (MON) & 200 & 40 \\
%Mycophenolic acid (MPA) & 200 & 40 \\
%Nivalenol (NIV) & 200 & 40 \\
%Zearalenone (ZEN) & 200 & 40 \\
%Beauvericin (BEA) & 100 & 20 \\
%DON-3-G (D3G) & 100 & 20 \\
%Enniatin A (ENN A) & 100 & 20 \\
%Enniatin A1 (ENN A1) & 100 & 20 \\
%Enniatin B (ENN B) & 100 & 20 \\
%Enniatin B1 (ENN B1) & 100 & 20 \\
%Patulin (PAT) & 100 & 20 \\
%Sterigmatocystin (STG) & 100 & 20 \\
%Ochratoxin A (OTA) & 10 & 2 \\
%Ochratoxin B (OTB) & 50 & 10 \\
%Aflatoxin B1 (AFB1) & 5 & 1 \\
%Aflatoxin B2 (AFB2) & 5 & 1 \\
%Aflatoxin G1 (AFG1) & 5 & 1 \\
%Aflatoxin G2 (AFG2) & 5 & 1 \\
%Aflatoxin M1 (AFM1) & 5 & 1 \\
%\bottomrule
%\end{tabular}
%\end{table}

\vspace{1em}

% Table 3
\begin{table}[H]
\centering
\caption{Details of the intra-day validation (intermediate precision), showing within-laboratory reproducibility ($RSD_{WLR}$), apparent recovery ($R_A$), and extraction efficiency ($R_E$). $R_E$ was calculated based on spiked samples versus spiked extracts at the high level (HL) on day 3.}
\begin{tabular}{@{}p{4.5cm} r r r r r@{}}
\toprule
Analyte & Mean Conc. & StdDev & $RSD_{WLR}$ & $R_A$ & $R_E$ \\
 & (\textmu g/kg) & & (\%) & (\%) & (\%) \\
\midrule
15-acetyl-deoxynivalenol & 114.4 & 15.1 & 13.2 & 57.2 & 95.2 \\
3-acetyl-deoxynivalenol & 190.4 & 5.0 & 2.7 & 95.2 & 99.0 \\
Alternariol & 164.3 & 16.1 & 9.8 & 82.2 & 91.7 \\
Alternariol methyl ether & 160.3 & 20.8 & 13.0 & 80.2 & 89.0 \\
Beauvericin & 93.8 & 5.1 & 5.4 & 93.8 & 100.9 \\
Deoxynivalenol & 191.2 & 6.2 & 3.2 & 95.6 & 100.8 \\
Deoxynivalenol-3-glucoside & 69.0 & 3.2 & 4.7 & 69.0 & 77.5 \\
Diacetoxyscirpenol & 187.0 & 10.0 & 5.3 & 93.5 & 99.5 \\
Enniatin A & 85.9 & 5.8 & 6.7 & 85.9 & 102.5 \\
Enniatin A1 & 88.8 & 4.8 & 5.4 & 88.8 & 98.3 \\
Enniatin B & 97.5 & 8.4 & 8.6 & 97.5 & 97.7 \\
Enniatin B1 & 93.8 & 6.7 & 7.2 & 93.8 & 98.2 \\
Ergocristine & 47.5 & 10.9 & 22.9 & 23.8 & 36.3 \\
Ergotamine & 124.5 & 11.3 & 9.1 & 62.2 & 93.4 \\
HT-2 toxin & 188.1 & 4.8 & 2.6 & 94.1 & 97.7 \\
Moniliformin & 173.6 & 11.1 & 6.4 & 86.8 & 74.0 \\
Nivalenol & 160.2 & 35.8 & 22.4 & 80.1 & 92.3 \\
Sterigmatocystin & 69.8 & 10.0 & 14.3 & 69.8 & 91.5 \\
T-2 toxin & 187.2 & 4.5 & 2.4 & 93.6 & 98.5 \\
T-2 toxin-3-glucoside & 175.3 & 7.9 & 4.5 & 87.6 & 100.9 \\
Zearalenone & 186.1 & 6.3 & 3.4 & 93.1 & 99.5 \\
\bottomrule
\end{tabular}
\end{table}

\vspace{1em}

% Table 4
\begin{table}[H]
\centering
\caption{Results of the Limit of Detection (LOD) and Limit of Quantification (LOQ) derived from low-level spikes, along with the expanded measurement uncertainty ($U$).}
\begin{tabular}{@{}p{4.5cm} r r r@{}}
\toprule
Analyte & LOD & LOQ & $U$ \\
 & (\textmu g/kg) & (\textmu g/kg) & (\%) \\
\midrule
15-acetyl-deoxynivalenol & 2.8 & 9.2 & 40 \\
3-acetyl-deoxynivalenol & 2.2 & 7.3 & 8 \\
Alternariol & 1.7 & 5.7 & 29 \\
Alternariol methyl ether & 1.3 & 4.2 & 39 \\
Beauvericin & 2.6 & 8.6 & 16 \\
Deoxynivalenol & 3.7 & 12.4 & 10 \\
Deoxynivalenol-3-glucoside & 1.7 & 5.5 & 14 \\
Diacetoxyscirpenol & 1.9 & 6.4 & 16 \\
Enniatin A & 1.0 & 3.4 & 20 \\
Enniatin A1 & 1.6 & 5.4 & 16 \\
Enniatin B & 5.1 & 17.0 & 26 \\
Enniatin B1 & 2.8 & 9.2 & 21 \\
Ergocristine & 1.6 & 5.2 & 69 \\
Ergotamine & 3.6 & 12.0 & 27 \\
HT-2 toxin & 4.1 & 13.6 & 8 \\
Moniliformin & 2.0 & 6.7 & 19 \\
Nivalenol & 19.0 & 63.2 & 67 \\
Sterigmatocystin & 0.8 & 2.7 & 43 \\
T-2 toxin & 3.4 & 11.4 & 7 \\
T-2 toxin-3-glucoside & 4.5 & 15.0 & 14 \\
Zearalenone & 1.8 & 6.1 & 10 \\
\bottomrule
\end{tabular}
\end{table}

% Table 4
%\begin{table}[H]
%\centering
%\caption{Results of the LOD and LOQ taken from the LL spikes, as well as the expanded measurement uncertainty (U).}
%\begin{tabular}{@{}l r r r@{}}
%\toprule
%Analyte & LOD (\textmu g/kg) & LOQ (\textmu g/kg) & U (\%) \\
%\midrule
%MON & 2.0 & 6.7 & 19\% \\
%15-AcDON & 2.8 & 9.2 & 40\% \\
%3-AcDON & 2.2 & 7.3 & 8\% \\
%AFB1 & 0.1 & 0.5 & 15\% \\
%AFB2 & 0.2 & 0.7 & 18\% \\
%AFG1 & 0.1 & 0.4 & 32\% \\
%AFG2 & 0.1 & 0.5 & 28\% \\
%AFM1 & 0.1 & 0.5 & 23\% \\
%Alternariol & 1.7 & 5.7 & 29\% \\
%AME & 1.3 & 4.2 & 39\% \\
%Beauvericin & 2.6 & 8.6 & 16\% \\
%DAS & 1.9 & 6.4 & 16\% \\
%DON & 3.7 & 12.4 & 10\% \\
%DON-3-G & 1.7 & 5.5 & 14\% \\
%EGT & 3.6 & 12.0 & 27\% \\
%Enniatin A & 1.0 & 3.4 & 20\% \\
%Enniatin A1 & 1.6 & 5.4 & 16\% \\
%Enniatin B & 5.1 & 17.0 & 26\% \\
%Enniatin B1 & 2.8 & 9.2 & 21\% \\
%Ergocornine & 2.3 & 7.6 & 51\% \\
%Ergocristine & 1.6 & 5.2 & 69\% \\
%FB1 & 4.2 & 13.8 & 25\% \\
%FB2 & 2.6 & 8.7 & 19\% \\
%HT-2 & 4.1 & 13.6 & 8\% \\
%MPA & 3.4 & 11.3 & 16\% \\
%Nivalenol & 19.0 & 63.2 & 67\% \\
%OTA & 0.3 & 1.0 & 13\% \\
%OTB & 0.7 & 2.4 & 12\% \\
%Patulin & 2.8 & 9.4 & 150\% \\
%Sterigmatocystin & 0.8 & 2.7 & 43\% \\
%T2G & 4.5 & 15.0 & 14\% \\
%T2-Toxin & 3.4 & 11.4 & 7\% \\
%ZEN & 1.8 & 6.1 & 10\% \\
%\bottomrule
%\end{tabular}
%\end{table}